\documentclass[lettersize,journal]{IEEEtran}
\usepackage{amsmath,amsfonts}
\usepackage{algorithmic}
\usepackage{algorithm}
\usepackage{array}
\usepackage[caption=false,font=normalsize,labelfont=sf,textfont=sf]{subfig}
\usepackage{textcomp}
\usepackage{stfloats}
\usepackage{url}
\usepackage{verbatim}
\usepackage{graphicx}
\usepackage{cite}
\usepackage{longtable}
\usepackage{supertabular}
\usepackage{multirow} 
\usepackage{caption}
\usepackage{float}
\usepackage{makecell}

\begin{document}
\bibliographystyle{plain}

\title{A novel framework for MCDM based on Z numbers and
soft likelihood function}

\author{Yuanpeng He
}

\markboth{Journal of \LaTeX\ Class Files,~Vol.~14, No.~8, August~2021}%
{Shell \MakeLowercase{\textit{et al.}}: A Sample Article Using IEEEtran.cls for IEEE Journals}


\maketitle

\begin{abstract}
The optimization on the structure of process of information management under uncertain environment has attracted lots of attention from researchers around the world. Nevertheless, how to obtain accurate and rational evaluation from assessments produced by experts is still an open problem. Specially, intuitionistic fuzzy set provides an effective solution in handling
indeterminate information. And Yager proposes a novel method for fusion of probabilistic evidence to handle uncertain and conflicting information lately which is called soft likelihood function. This paper devises a novel framework of soft likelihood function based on information volume of fuzzy membership and credibility measure for extracting truly useful and valuable information from uncertainty. An application is provided to verify the validity and correctness of the proposed framework. Besides, the comparisons with other existing methods further demonstrate the superiority of the novel framework of soft likelihood function.
\end{abstract}


\section{Introduction}
\IEEEPARstart{D}{ue} to its effectiveness in handling uncertain information, multi-criterion applied in lots of relative fields, including but not limited to risk evaluation \cite{ref1,ref2,ref3}, supplier selection \cite{ref4,ref5,ref6}, fault diagnosis \cite{ref7,ref8}. However, how to obtain accurate decisions from uncertain environment and complex source is still an open issue \cite{he2021conflicting, he2022mmget}.

Traditionally, experts are supposed to provide accurate evaluation on actual situations based on random alternative \cite{ref9}. But the process of producing judgments may be too arbitrary and subjective, which can lead to unexpected results. Therefore, a new concept of linguistic variable is designed to better conform to intuition of human instead of assigning specifically certain values to subjects. In other words, it is not reasonable to require experts to provide accurate judgments on different situations \cite{he2024matrixbaseddistancepythagoreanfuzzy}. Besides, it can be concluded that the introduction of linguistic variable enables decision making to become more feasible \cite{ref10,ref11,ref12}. Utilizing linguistic variables, some useful and meaningful techniques are developed such as intuitionistic fuzzy set (IFS) \cite{ref13,ref14,ref15,ref16}, interval-valued IFS \cite{ref17,ref18,ref19} and other categories of fuzzy technologies \cite{ref20,ref21,ref22, he2023tdqmf}. 

In real life, in the process of generating estimations, experts may encounter different circumstances which shape varied judgments on actual conditions \cite{li2022nndf, he2024residual, xu2023spatio}. Degree of affirmation, negation and hesitancy from experts change with standard of knowledge or other potential influential factors in the process of recognition of situations. Compared with relative approaches in information modeling, IFS is more flexible and possesses superiority in handling uncertainty and conflicts.

Furthermore, the Dempster-Shafer (D-S) evidence theory has been recognized as an efficient framework to handle uncertainty and incomplete information \cite{he2022new, he2023ordinal}. It offers a robust mechanism for combining evidence from different sources to arrive at a degree of belief that considers all available information. D-S evidence theory extends the classical probability theory by allowing the representation of uncertainty and conflict explicitly, making it particularly effective in multi-criteria decision-making scenarios. The integration of D-S evidence theory into decision-making processes enhances the ability to deal with complex and uncertain environments, offering a complementary perspective to existing fuzzy technologies and improving overall robustness \cite{he2022ordinal}.

Moreover, in order to appropriately and effectively manage ambiguity, Yager propose a softer process in combining uncertain information \cite{ref23} based on original likelihood function and OWA operators \cite{ref24} to avoid generating too absolute and counter-intuitive results, which is named as soft likelihood function (SLF). And some interesting researches are developed on the base of SLF \cite{ref25,ref26,ref27} which provide some efficient solutions to extract instructive information from uncertainty. Because of the feasibility
and excellent performance of soft likelihood function \cite{he2024generalized}, it is introduced into the process of MCDM under intuitionistic environment in order to obtain correct and reasonable proofs for decision and strategy designing in this paper. And the contributions of the proposed framework for MCDM are listed as follows:
\begin{itemize}
\item Information volume of fuzzy membership and credibility measure of judgments of experts are properly combined to serve as a new dimension of estimation of each judgment among the ones belongs to groups.
\item A novel method of producing fuzzy preference relation is devised for detecting degrees of importance of every node contained in information, which further improves sensitivity of the whole framework in erasing dirty data.
\item A process of generating varied attitude characters (ACs) is designed. The superiority of this operation is that the phenomenon of setting AC for information manually and subjectively is avoided, which reduces the possibility in producing counter-intuitive results on account of unexpected fictitious factors.
\item The proposed framework for SLF is able to obtain sufficiently accurate and rational results for decision and strategy making.
\end{itemize}
The remaining part of the paper is organized as follows. The section of
preliminaries generally introduces relevant information of useful concepts of techniques. And in the next section, some necessary discussions about the proposed framework are provided. Besides, detailed process of method proposed is well illustrated in the forth section. Moreover, an application and corresponding analysis are given to verify the correctness and validity of the proposed framework. In the last, conclusions are made to summarize the value of the work presented in this paper.

\section{ Preliminaries}
Some related concepts are briefly introduced in this part. Some interesting work are completed based on them, such as decision making under intuitionistic environment \cite{ref28,ref29}, network model \cite{ref30}, expert decision using Z-number \cite{ref31} and information management based on soft likelihood function \cite{ref32}.

\subsection{Intuitionistic fuzzy set \cite{ref13} }
Let $R$ be a finite universe of discourse. Then, an IFS $T$ can be defined as:
\begin{equation}
T = \{\langle r, \varrho_T(r), \varphi_T(r) \rangle \mid r \in R \} \quad (1)
\end{equation}

where
\begin{equation}
\varrho_T(r) : R \rightarrow [0, 1], \varphi_T(r) : R \rightarrow [0, 1] \quad (2)
\end{equation}

and the property satisfied by the IFS can be defined as:
\begin{equation}
0 \leq \varrho_T(r) + \varphi_T(r) \leq 1, \ \forall r \in R \quad (3)
\end{equation}

in which $\varrho_T(r)$ indicates the degree of membership and $\varphi_T(r)$ indicates the degree of non-membership with respect to $r \in R$.

Besides, the degree of hesitance can be defined as:
\begin{equation}
\xi_T(r) = 1 - \varrho_T(r) - \varphi_T(r) \quad (4)
\end{equation}

Moreover, in a more simple expression, the tuple $t=(\varrho_T(r), \varphi_T(r),  \xi_T(r))$ is utilized to represent an intuitionistic fuzzy number (IFN), which can be regarded as one element of set $T$.

\subsection{First-order information volume of fuzzy membership function \cite{ref33}}
Assume there exist one IFN $t=(\varrho_T(r), \varphi_T(r),  \xi_T(r))$, then the information volume of it can be defined as:
\begin{align}
EIFN(t) = -(\varrho_T(r)\log_2\varrho_T(r) + & \varphi_T(r)\log_2\varphi_T(r) \ + \nonumber\\
& \xi_T(r)\log_2 \frac{\xi_T(r)}{C})\tag{5}
\end{align}

in which the parameter $C$ indicates the cardinal number of fuzzy sets which can be defined as:
\begin{equation}
 C=\begin{cases} 2& \text{(traditional f uzzy sets)} \\
  3& \text{(intuitionistic f uzzy set)}
\end{cases}\tag{6}
\end{equation}

After evaluation, the value of entropy reaches its zenith when $\varrho_T(r) = \varphi_T(r) = \frac{1}{5}$, $\xi_T(r) = \frac{3}{5}$.

\subsection{Score function for IFS}
Assume there exist one IFN $t=(\varrho_T(r), \varphi_T(r),  \xi_T(r))$, then the score
function can be defined as:
\begin{equation}
    SF(t) = \frac{1}{2}(1+\xi_T)(1-\varrho_T)\tag{7}
\end{equation}

It can be concluded that when the value of $SF(t)$ becomes smaller, $T(r)$ is regarded less reliable.

\subsection{A Jensen–Shannon divergence-based distance measure for IFS \cite{ref34}}

Assume there exist two IFSs $G = \{\langle r,\varrho_G(r), \varphi_G(r) \rangle \mid r \in R\}$ and $V = \{<r,\varrho_V(r), \varphi_V(r) > \mid r \in R\}$ under the finite universe of discourse $R$. And
the degree of hesitancy can be obtained by $\xi_G(r) = 1 -  \varrho_G(r) - \varphi_G(r)$ and $\xi_V(r) = 1 - \varrho_V(r) - \varphi_V(r)$. Then, the distance measure of IFS can be defined as:
\begin{align}
\begin{aligned}
&D(Z,V) =\\
&[\frac{1}{2}(\varrho_{G}(r)log_{2}\frac{2*\varrho_{G}(r)}{\varrho_{G}(r)+\varrho_{V}(r)}+\varrho_{V}(r)log_{2}\frac{2*\varrho_{V}(r)}{\varrho_{G}(r)+\varrho_{V}(r)}  \\
&+\varphi_{\mathrm{G}}(r)log_{2}\frac{2*\varphi_{G}(r)}{\varphi_{\mathrm{G}}(r)+\varphi_{V}(r)}+\varphi_{V}(r)log_{2}\frac{2*\varphi_{V}(r)}{\varphi_{G}(r)+\varphi_{V}(r)} \\
&+\xi_{G}(r)log_{2}\frac{2*\xi_{G}(r)}{\xi_{G}(r)+\xi_{V}(r)}+\xi_{V}(r)log_{2}\frac{2*\xi_{V}(r)}{\xi_{G}(r)+\xi_{V}(r)})]^{\frac{1}{2}}
\end{aligned}\tag{8}
\end{align}

\subsection{Closeness centrality \cite{ref35}}
The closeness centrality represents the level of difficulty of the process
that one point reaches other points which can be defined as:
\begin{equation}
    CC_f = \frac{\vert F \vert -1}{\sum_{f \neq j} D_{fj}}\tag{9}
\end{equation}

in which $ \vert F \vert $ indicates the total number of points and $Df_j$ represents the distance between $f$ and $j$.

\subsection{ Z-number \cite{ref36}}
Z-number is proposed by Zadeh to better handle uncertain information
which can be defined as:
\begin{equation}
    ZN=\{\Upsilon, U \} = ZN^+(\Upsilon,\tau_\Upsilon \cdot p_{H_\Upsilon} \text{ is }U)\tag{10}
\end{equation}

$\Upsilon $ is a fuzzy constraint with respect to certain problem $H_\Upsilon$ and $U$ is an
estimation of reliability of $\Upsilon $. Besides, the variable $H_\Upsilon$ is stochastic for $\Upsilon $ and the membership of $\Upsilon $ and $U$ are represented by $\varrho_\Upsilon(r) $ and $\varrho_U(r)$. Additionally, $\Upsilon $ and $U$ are not independent with each other which are connected by the hidden probability $p_{H_\Upsilon}$.

\subsection {Fuzzy preference relation \cite{ref37}}
Assume there exist a set $T=\{t_1, t_2,\ldots, t_k \}$, a fuzzy preference relation F can be expressed by a complementary matrix FPRM = (fwe)k×k which can be defined as:
\begin{equation}
f_{w e}>0, f_{w e}+f_{e w}=1, f_{w w}=1, w, e=1,2, \ldots, k \tag{11}
\end{equation}

where the $f_{w e}$ represents the extent that $t_w$ is preferred to $t_e$. And the details of parameter $f_{w e}$ can be defined as:
\begin{equation}
f_{w e}=\left\{\begin{array}{ll}
1 & t_{w} \text { is strongly preferred to } t_{e} \\
\vartheta \in(0.5,1) & t_{w} \text { is slightly preferred to } t_{e} \\
0.5 & \text { no difference between } t_{w} \text { and } t_{e} \\
\omega \in(0,0.5) & t_{e} \text { is slightly preferred to } t_{w} \\
0 & t_{e} \text { is strongly preferred to } t_{w}
\end{array}\right. \tag{12}
\end{equation}
in which the bigger the value of $\vartheta$ the more degree of preference whit respect to $t_w$ to $t_e$. On the contrary, the value of $\omega$ becomes less, more preference is given to $t_e$.

\subsection{Original likelihood function}
Assume there exist a series of distributions of probability, $dp_j$, in regard
to incident $I_x$ and the according likelihood function which can be defined as:
\begin{equation}
    LF_{I_x}=\prod_{j=1}^kdp_j  \tag{13}
\end{equation}

\subsection{Ordered weighted aggregation operator \cite{ref24}}
The ordered weighted aggregation operator is composed of $k$ dimensions and is a mapping of $Wk \rightarrow  W $which can be defined as:
\begin{equation}
\psi=\begin{bmatrix}w_1\\w_2\\\ldots\\w_k\end{bmatrix}\tag{14}
\end{equation}

and the property of the operator can be defined as:
\begin{equation}
    \sum_{j=1}^kw_j=1,0\leq w_j\leq1 \tag{15}
\end{equation}

When $\psi=\psi^*=[1,0,...,0]^T$, it can be regarded as a optimized strategy;when$\psi=\psi^*=[0,0,...,1]^T$, it can be regarded as a pessimistic strategy.Besides, if$\psi=\psi_K=[\frac1k,\frac1{\bar{k}},...,\frac1{\bar{k}}]^T$. Based on the regulations, an attitude
character $\Delta$ is devised and then one method to produce weights for OWA operator can be defined as:
\begin{equation}
    w_j=(\frac sk)^{\frac{1-\Delta}\Delta}-(\frac{s-1}k)^{\frac{1-\Delta}\Delta} \tag{16}
\end{equation}

\subsection{Soft likelihood function \cite{ref23}}
Yager thinks the expression of likelihood function is too absolute, so a softer process of creating likelihood function is designed based on the definition of OWA operators which can be defined as:
\begin{equation}
    \Upsilon S_l(a)=\subseteq_{h=1}^adp_l\iota_l(h) \tag{17}
\end{equation}

Utilizing OWA operators, the formula for soft likelihood function $E_l$ in regard to one element $c_l$ can be defined as:
\begin{equation}
    E_{l,\psi}=\sum_{a=1}^kw_a \Upsilon S_l(a) \tag{18}
\end{equation}

\section{Some discussions on the design and performance of the presented method}
The method proposed in this paper first takes the relationship of individual IFSs which are within one group of judgments made by one expert
to further indicate the role of each part of a group of IFS. Then, by considering the situation of the judgments in all estimation produce by expert, the
total degree of reliability of judgments can be obtained. Besides, another
dimension of assessing the conditions of judgments in the form of IFS individually is designed by introducing the information volume of fuzzy membership function to ensure the judgments are assigned a comprehensive
estimation. Through combining the weights generated by reliability measure and information volume, a new framework based on dynamic attitude
characters of soft likelihood function can be obtained. Compared with the
traditional formula of soft likelihood function \cite{ref23}, the difference is that the
attitude characters in the traditional one are set subjectively without taking
internal and external factors of these judgments into consideration, which
is not a intuitive operation because the attitude character can not be exactly
the same for different judgments. However, in some related works, the attitude character dose not vary with the changes of source of information and
conditions of information itself. In order to overcome this drawback, the
method of generating attitude character is designed and the corresponding
formula of varied expression of improved soft likelihood function is also
produced. After necessary checking and comparisons with other effective
method, the improved version of soft likelihood function performs very
well in decision making and strategy designing.

Besides, the expectation of the performance of the presented method
can be concluded into two points. The first one is that the proposed method
highlights the most credible judgments produced by experts and exaggerates the values of indicators with respect to estimations on actual situations.
Therefore, the differences of values of indicator can be obvious enough and
even very divergent with each other, which is determined by the feature
of the method of generation of attitude character and original soft likelihood function. The second is an extension of the first one. The crucial
point needing to be emphasized is that what is expected to be concerned
is the final results of the judgments on certain subjects instead of focusing
on the difference of values of incompletely processed data. For decision
makers, the thing to determine the priority of different decisions and strategy is the most important. Therefore, the values of judgments produced by
algorithms are expected to be straightforward and clear for decision and
strategy making, then the performance and the credibility of them can be
believable and efficient


\section{Novel framework of soft likelihood function in expert decision systems}
In this section, a new framework of soft likelihood function is proposed to handle multiple criteria decision making problems. The presented
method takes attributes of information provided into consideration fully
so as to generate enough accurate estimations to actual situations. And the
detailed process of the presented method is given as below:

\textbf{Step 1}: Obtain corresponding information for linguistic variables in decision making. Then, transfer them into the form of IFS. Linguistic variables
are widely utilized to manage uncertain information for decision makers.
One thing which is supposed to be pointed out is that if experts are familiar
with the concept of IFS, they can make judgments directly based on IFS.

\textbf{Step 2}: Utilize the concept of traditional score function, a formula is
devised to calculate measurement of reliability of each individual IFS. For
example, assume there exist one IFS $I=(\varrho_T(r),\varphi_T(r))$, the degree of reliability of it can be defined as:
\begin{equation}
    SF(t)^I=\frac12(1+\varrho_T)(1-\varphi_T)  \tag{19}
\end{equation}

Because the degree of membership is a crucial factor which indicates the
general situation of judgments given by experts, then the formula focuses
on the influence brought by the membership and the part which dose not
opposes certain subjects evaluated.

\textbf{Step 3}: Then, the fuzzy numbers can be transferred into the form which
can be defined as:
\begin{equation}
    I_Z=((\varrho_T(r),\varphi_T(r)),SF(t)^I) \tag{20}
\end{equation}

\textbf{Step 4}: Combine the transformed fuzzy numbers by getting products of
degree of reliability and membership and non-membership. The detailed
process can be defined as:
\begin{equation}
    I=(\varrho_T(r)\times SF(t)^I,\varphi_T(r)\times SF(t)^I)\tag{21}
\end{equation}

\textbf{Step 5}: Calculate distances of IFSs within groups, $D(Z, V)$, by utilizing
Jensen–Shannon divergence-based distance measure which can be defined as:
\begin{equation}
\begin{aligned}
&D(Z,V) =\\
&[\frac12(\varrho_{Z}(r)log_{2}\frac{2*\varrho_{Z}(r)}{\varrho_{Z}(r)+\varrho_{V}(r)}+\varrho_{V}(r)log_{2}\frac{2*\varrho_{V}(r)}{\varrho_{Z}(r)+\varrho_{V}(r)}  \\
&+\varphi_Z(r)log_2\frac{2*\varphi_Z(r)}{\varphi_Z(r)+\varphi_V(r)}+\varphi_V(r)log_2\frac{2*\varphi_V(r)}{\varphi_Z(r)+\varphi_V(r)} \\
&+\xi_{Z}(r)log_{2}\frac{2*\xi_{Z}(r)}{\xi_{Z}(r)+\xi_{V}(r)}+\xi_{V}(r)log_{2}\frac{2*\xi_{V}(r)}{\xi_{Z}(r)+\xi_{V}(r)})]^{\frac12}  
\end{aligned}\tag{22}
\end{equation}


\textbf{Step 6}: Produce similarity measure for IFSs. Assume there exist one IFS
$Z_i$ and other IFSs which are contained in one group with $Z_i$ are denoted by $Z_k$ and the total number of IFSs in one group is $M$. Then, the similarity for
IFS $Z_i$ can be produced by the formula which can be defined as:
\begin{equation}
SM_{Z_i}=\frac{M-1}{\sum_{k\neq i}D(Z_i,Z_k)} \tag{23}
\end{equation}

\textbf{Step 7}: Generate fuzzy preference relation in groups of IFSs. Assume
there exist two similarity of IFSs which can be given as $SM_{Z_i}$ and $SM_{Z_j}$.
Then, normalize the two similarities using the formula which can be de-
fined as:
\begin{equation}
\vartheta=\frac{SM_{Z_i}}{SM_{Z_i}+SM_{Z_j}},\varphi=\frac{SM_{Z_j}}{SM_{Z_i}+SM_{Z_j}}\tag{24}
\end{equation} 
then, compare the two parameter $\vartheta$ and $\omega$, if $\vartheta$ is bigger than $\omega$, then it can be regarded that $Z_j$ prefers $Z_i$ to be considered as the reference in the decision making; on the contrary, if $\omega$ is bigger than $\vartheta$, then it can be regarded that $Z_i$ prefers $Z_j$ to be considered as the reference in the decision making. Besides, when $\vartheta$ and $\omega$ are identically equal, there exist no winner in comparison. The preferred ones are allocated an indicator of 1 and the other ones are distributed 0, then one fuzzy preference relation matrix can be constructed which can be defined as:
\begin{equation}
FPRM=\begin{bmatrix}pr_{11}&pr_{12}&\cdots&pr_{1k}\\pr_{21}&pr_{22}&\cdots&pr_{2k}\\\vdots&\vdots&\cdots&\vdots\\pr_{k1}&pr_{k2}&\cdots&pr_{kk}\end{bmatrix},pr_{ij}=0,1  \tag{25}
\end{equation}

The final results of FPRM with respect to one IFS are summed to indicate the importance of it in decision making. For one IFS $Z_i$, its corresponding point can be defined as:
\begin{equation}
PO_{Z_i}=\sum_{j\neq i}pr_{ij} \tag{26}
\end{equation}

\textbf{Step 8}: Obtain the final judgments of IFSs withing groups using the results of points obtained based on the fuzzy preference relation. Assume there exist one groups of judgments in the form of IFS and the according values of points can be given as $A_{Z_i}=\{PO_{Z_1},PO_{Z_2},...,PO_{Z_k}\}.$ Then, the weights indicating the role of calculating divergences among groups of IFSs for any one of IFSs can be defined as:
\begin{equation}
B_{Z_i}=\frac{PO_{Z_i}}{\sum_{j=1}^kPO_{Z_j}}    \tag{27}
\end{equation}

\textbf{Step 9}:Calculate total distances of groups of IFSs using judgments. Assume there exist the first group of IFSs which is denoted by $Z_i^1$ and othen groups which are represented by $\{V_i^2,...,V_i^N\}$. Then, the total distance between groups of $Z_i^1$ and $V_i^2$ can be calculated as:
\begin{equation}
D(Z_i^1,V_i^2)^{total}=\sum_{i=1}^kB_{Z_i}\times D(Z_i^1,V_i^2) \tag{28}   
\end{equation}

\textbf{Step 10}: Compute the credibility measurement of groups of IFSs. Taking the complete environment of groups of IFSs within judgments produced by one expert into consideration, the degree of credibility of one
group $Z_i^1$ can be defined as:
\begin{equation}
CR_{Z_i^1}=D(Z_i^1,V_i^2)_{max}^{total}-D(Z_i^1,V_i^2)^{total} \tag{29}
\end{equation}

\textbf{Step 11}: Normalize the degree of credibility. And the process of the
normalization with respect to $Z_i^1$ can be defined as:
\begin{equation}
CR_{Z_i^1}^{Nor}=\frac{CR_{Z_i^1}}{CR_{Z_i^1}+\sum_{j=2}^NCR_{V_i^j}} \tag{30}
\end{equation}

\textbf{Step 12}: Obtain the total information volume of each groups of original IFSs. Utilizing the first-order information volume of fuzzy membership
function, the corresponding information volume with respect to $Z_i^1$ can be calculated as:
\begin{equation}
\begin{aligned}
IVF_{Z_i^1}=\sum_{i=1}^k-(\varrho_T(r)^{Z_i^1}log_2  \varrho_T(r)^{Z_i^1} &+ \varphi_T(r)^{Z_i^1}
log_2\varphi_T(r)^{Z_i^1}\\+
&\xi_T(r)^{Z_i^1}log_2\frac{\xi_T(r)^{Z_i^1}}{C})
\end{aligned}\tag{31}
\end{equation}

\textbf{Step 13}: Generate the modified values of information volume to avoid producing a weights which is exactly equal to 0. And the process of it in regard to $Z_i^1$ can be defined as:
\begin{equation}
IVF_{Z_i^1}^M=e^{IVF_{Z_i^1}} \tag{32}
\end{equation}

\textbf{Step 14}: Normalize the modified values of information volume to adapt to process of generating attitude character, the process with respect to $Z_i^1$ can be defined as:
\begin{equation}
IVF_{Z_i^1}^{MN}=\frac{IVF_{Z_i^1}^M}{IVF_{Z_i^1}^M+\sum_{j=2}^NIVF_{V_i^j}^M} \tag{33}
\end{equation}

\textbf{Step 15}: Generate attitude character utilizing degree of credibility and modified information volume. The process about $Z_i^1$ can be defined as:
\begin{equation}
\alpha_{Z_i^1}=IVF_{Z_i^1}^{MN}\times CR_{Z_i^1}^{Nor}
\tag{34}
\end{equation}

\textbf{Step 16}: Normalize the attitude character to be adaptive to construction of soft likelihood function. Besides, the corresponding formula can be
defined as:
\begin{equation}
\alpha_{Z_i^1}^{Nor}=\frac{\alpha_{Z_i^1}}{\alpha_{Z_i^1}+\sum_{j=2}^N\alpha_{V_i^j}} \tag{35}
\end{equation}

\textbf{Step 17}: According to the attitude characters obtained, the corresponding formula of soft likelihood function about $Z_i^1$ can be designed which can be defined as:
\begin{equation}
P=\frac{1-\alpha_{Z_i^1}^{Nor}}{\alpha_{Z_i^1}^{Nor}} \tag{36}
\end{equation}
\begin{equation}
\left.DSLF_{Z_i^1}=\left\{\begin{array}{c}\sum_{j=1}^k((\frac{j}{k})^P-(\frac{j-1}{k})^P)\prod_{h=1}^kw_h\Upsilon S_l(h)\end{array}\right.\right.    \tag{37}
\end{equation}

What is expected to be pointed out is that $w_h$ and $\Upsilon S_l(h)$ are calculated under the situation of $Z_i^1$.

\textbf{Step 18}: Input the attitude characters obtained into the formula of soft
likelihood function to generate corresponding judgments with respect to each group of IFSs.

\textbf{Step 19}: For every IFS in the estimation made by experts, all of them
is distributed to a sequence to adapt to demand of soft likelihood function.
For example, the IFSs in the column of x1 are allocated a sequence of 1 and
the other ones in the column x2 are distributed a order of 2. With respect to
other ones, the method of assigning sequences are in the same manner.

\textbf{Step 20}: Assume the IFSs in one group can be denoted by $A^Modified = (\varrho_T(r)^i,\varphi_T(r)^i)$.
Then, for the process of iteration of construction of basic likelihood function, the element dpi can be replaced by $\varrho_T(r)^i-\varphi_T(r)^i$ which can be defined as:
\begin{equation}
dp_i=\varrho_T(r)^i-\varphi_T(r)^i    \tag{38}
\end{equation}

Besides, the values in each group is supposed to be sorted.

\textbf{Step 21}: Suppose the judgments of experts can be divided into three
groups of IFSs which can be denoted by $\{Z_i^1, V_i^2, V_i
^2\}$. Then, the gross estimation of the suggestions made by the expert can be defined as:
\begin{equation}
GE_{Expert}=\frac{DSLF_{Z_i^1}+DSLF_{V_i^2}+DSLF_{V_i^3}}{0.03} \tag{39}    
\end{equation}

The value of the final judgments are amplified 100 times to ensure the
process of comparison of these values becomes more convenient.

\textbf{Step 22}: Rank the values of gross estimation, $GEE_xpert$, with respect to
different subjects.

\textbf{Step 23}: Obtain the processed information and description of actual
conditions.

\textbf{Note}:In \textbf{Step 6-21}, the corresponding values of $\{V_i^2, \dots, V_i^N\}$ can be calculated in the same manner as $Z_i^1$. Moreover, the detailed process of the presented method is given in Figure \ref{fig_1}.

\begin{table*}[ht]
\renewcommand{\arraystretch}{1.0}
\setlength{\tabcolsep}{10pt}
\centering
\begin{tabular}{c c c c c c c c c}
\hline
Round & Alternative  & & x1 & x2 & x3 & x4 &x5 &x6 \\
\hline
r=1 & $Supplier_1$ & $Expert_1$ & (0.6,0.2) & (0.3,0.5) & (0.7,0.1) & (0.6,0.1) &(0.5,0.3) &-\\
& &$Expert_2$ & (0.3,0.4) & (0.4,0.4) & (0.5,0.2) & (0.7,0.2) & (0.5,0.3) &-\\
& &$Expert_3$ & (0.5,0.4) & (0.4,0.3) & (0.6,0.1) & (0.5,0.3) & (0.3,0.4) &-\\

& $Supplier_2$ & $Expert_1$ & (0.8,0.0) & (0.3,0.6) & (0.5,0.4) & (0.3,0.5) &(0.6,0.2) &-\\
& &$Expert_2$ & (0.7,0.3) & (0.4,0.5) & (0.4,0.3) & (0.6,0.1) & (0.5,0.2) &-\\
& &$Expert_3$ & (0.8,0.1) & (0.6,0.2) & (0.4,0.5) & (0.4,0.4) & (0.6,0.1) &-\\

& $Supplier_3$ & $Expert_1$ & (0.3,0.6) & (0.7,0.0) & (0.6,0.2) & (0.5,0.3) &(0.6,0.2) &-\\
& &$Expert_2$ & (0.5,0.2) & (0.6,0.3) & (0.5,0.1) & (0.4,0.4) & (0.8,0.2) &-\\
& &$Expert_3$ & (0.7,0.2) & (0.5,0.2) & (0.6,0.3) & (0.9,0.1) & (0.7,0.1) &-\\

& $Supplier_4$ & $Expert_1$ & (0.6,0.3) & (0.7,0.2) & (0.6,0.2) & (0.7,0.1) &(0.7,0.1) &-\\
& &$Expert_2$ & (0.7,0.1) & (0.7,0.2) & (0.5,0.2) & (0.7,0.1) & (0.7,0.1) &-\\
& &$Expert_3$ & (0.7,0.1) & (0.7,0.2) & (0.8,0.1) & (0.7,0.1) & (0.7,0.2) &-\\

& $Supplier_5$ & $Expert_1$ & (0.4,0.5) & (0.6,0.2) & (0.3,0.5) & (0.3,0.5) &(0.4,0.3) &-\\
& &$Expert_2$ & (0.4,0.4) & (0.5,0.3) & (0.2,0.3) & (0.5,0.3) & (0.6,0.2) &-\\
& &$Expert_3$ & (0.6,0.2) & (0.5,0.3) & (0.6,0.2) & (0.6,0.2) & (0.5,0.3) &-\\
\hline

r=2& $Supplier_1$ & $Expert_1$ & (0.7,0.1) & (0.4,0.4) & (0.5,0.1) & (0.5,0.1) &(0.4,0.4) &(0.6,0.3)\\
& &$Expert_2$ & (0.8,0.1) & (0.5,0.3) & (0.6,0.2) & (0.5,0.4) & (0.5,0.3) &(0.5,0.4)\\
& &$Expert_3$ & (0.3,0.4) & (0.5,0.3) & (0.6,0.1) & (0.5,0.3) & (0.2,0.4) &(0.6,0.2)\\

& $Supplier_2$ & $Expert_1$ & (0.8,0.1) & (0.5,0.3) & (0.6,0.2) & (0.5,0.4) &(0.5,0.3) &(0.5,0.2)\\
& &$Expert_2$ & (0.6,0.2) & (0.5,0.4) & (0.5,0.3) & (0.6,0.1) & (0.6,0.2) &(0.6,0.1)\\
& &$Expert_3$ & (0.5,0.2) & (0.4,0.2) & (0.5,0.4) & (0.5,0.4) & (0.6,0.1) &(0.7,0.1)\\

& $Supplier_4$ & $Expert_1$ & (0.7,0.3) & (0.6,0.3) & (0.6,0.2) & (0.7,0.1) &(0.6,0.1) &(0.7,0.1)\\
& &$Expert_2$ & (0.7,0.1) & (0.6,0.2) & (0.5,0.2) & (0.6,0.1) & (0.8,0.1) &(0.6,0.2)\\
& &$Expert_3$ & (0.8,0.1) & (0.6,0.3) & (0.5,0.1) & (0.8,0.1) & (0.7,0.2) &(0.7,0.2)\\

& $Supplier_5$ & $Expert_1$ & (0.5,0.3) & (0.6,0.2) & (0.5,0.1) & (0.4,0.4) &(0.3,0.3) &(0.7,0.2)\\
& &$Expert_2$ & (0.5,0.3) & (0.6,0.1) & (0.4,0.3) & (0.5,0.2) & (0.5,0.2) &(0.7,0.2)\\
& &$Expert_3$ & (0.6,0.1) & (0.4,0.4) & (0.6,0.2) & (0.5,0.2) & (0.5,0.3) &(0.6,0.2)\\

\hline
& $Supplier_1$ & $Expert_1$ & (0.6,0.2) & (0.5,0.3) & (0.6,0.1) & (0.6,0.2) &(0.5,0.1) &-\\
& &$Expert_2$ & (0.5,0.2) & (0.5,0.2) & (0.6,0.3) & (0.4,0.2) & (0.5,0.3) &-\\
& &$Expert_3$ & (0.5,0.4) & (0.4,0.3) & (0.5,0.1) & (0.6,0.3) & (0.4,0.4) &-\\

& $Supplier_2$ & $Expert_1$ & (0.7,0.2) & (0.6,0.2) & (0.5,0.1) & (0.5,0.4) &(0.5,0.2) &-\\
& &$Expert_2$ & (0.6,0.1) & (0.6,0.1) & (0.7,0.2) & (0.5,0.1) & (0.6,0.2) &-\\
& &$Expert_3$ & (0.6,0.2) & (0.5,0.2) & (0.5,0.3) & (0.5,0.4) & (0.6,0.2) &-\\

& $Supplier_3$ & $Expert_1$ & (0.7,0.1) & (0.6,0.3) & (0.6,0.2) & (0.7,0.1) &(0.6,0.1) &-\\
& &$Expert_2$ & (0.6,0.2) & (0.7,0.2) & (0.6,0.1) & (0.6,0.1) & (0.8,0.1) &-\\
& &$Expert_3$ & (0.7,0.1) & (0.7,0.1) & (0.6,0.2) & (0.7,0.1) & (0.5,0.2) &-\\

& $Supplier_4$ & $Expert_1$ & (0.8,0.1) & (0.7,0.2) & (0.6,0.1) & (0.7,0.2) &(0.7,0.1) &-\\
& &$Expert_2$ & (0.8,0.1) & (0.6,0.1) & (0.7,0.2) & (0.7,0.2) & (0.6,0.2) &-\\
& &$Expert_3$ & (0.7,0.1) & (0.8,0.1) & (0.7,0.1) & (0.7,0.2) & (0.6,0.3) &-\\

& $Supplier_5$ & $Expert_1$ & (0.6,0.3) & (0.7,0.2) & (0.6,0.1) & (0.8,0.1) &(0.6,0.2) &-\\
& &$Expert_2$ & (0.7,0.2) & (0.6,0.1) & (0.8,0.1) & (0.5,0.1) & (0.6,0.3) &-\\
& &$Expert_3$ & (0.6,0.3) & (0.5,0.4) & (0.6,0.2) & (0.7,0.1) & (0.8,0.1) &-\\

& $Supplier_6$ & $Expert_1$ & (0.7,0.2) & (0.8,0.1) & (0.7,0.2) & (0.8,0.1) &(0.7,0.2) &-\\
& &$Expert_2$ & (0.7,0.2) & (0.6,0.1) & (0.8,0.1) & (0.6,0.3) & (0.7,0.1) &-\\
& &$Expert_3$ & (0.8,0.1) & (0.7,0.2) & (0.6,0.2) & (0.6,0.3) & (0.8,0.1) &-\\
\hline

\end{tabular}
\caption{Detailed information about the judgments from experts}
\label{tab1}
\end{table*}

\section{Application}
In this section, an application is provided to illustrate the efficiency of the presented method.

Nowadays, in general, manufactures may establish relationships with material suppliers to make sure the sources of materials of production of
products are abundant and healthy. However, it is necessary to make estimation about the quality of materials from suppliers from time to time, because the environment of supply and relation in collaboration may vary,
which indicates that it is a metabolic problem. Moreover, due to the complexity of the problem, how to fully utilize uncertain information offered is one of the most important concern in fuzzy decision making under this
kind of case. Therefore, the proposed algorithm in handling uncertain information possesses superiority in this filed, because it is exactly designed to sufficiently extract all useful part of information to serve as a proof in
decision and strategy making.
\begin{table*}[htb]
\renewcommand{\arraystretch}{1.4}
\setlength{\tabcolsep}{4pt}
\centering
\begin{tabular}{c c c c c}
\hline
& AQM[38]  &Static ranking[39] &Dynamic ranking[39] &Presented method \\
$Round_1$ &$S_3>S_4>S_2>S_1>S_5$ &$S_4>S_3>S_2>S_1>S_5$ &$S_4>S_3>S_2>S_1>S_5$ &$S_4>S_3>S_2>S_1>S_5$\\
$Round_2$ &$S_4>S_2>S_5>S_1$ &$S_4>S_2>S_5>S_1$ &$S_4>S_2>S_5>S_1$ &$S_4>S_2>S_5>S_1$\\
$Round_3$ &$S_6>S_4>S_3>S_5>S_2>S_1$ &$S_6>S_4>S_5>S_3>S_2>S_1$ &$S_6>S_4>S_3>S_5>S_2>S_1$ &$S_6>S_4>S_3>S_5>S_2>S_1$\\
\hline
\end{tabular}
\caption{Results obtained by four methods (Note: Supplier is denoted by S)}
\label{result}
\end{table*}

Assume there exist one review meeting and three experts participate it to make assessment on five suppliers. More specifically, in order to make the process of estimation on the suppliers more accurate under the condition that $Supplier_3$ quits the process of estimation, each expert appends one more vote in the second round. Besides, in the round three, a new $Supplier_6$ joins in the review meeting. And the detailed information about the judgments in provided in Table \ref{tab1}.

Based on the description of the problem, the process of proposed method is divided into three parts, namely round 1, 2 and 3. The calculation of the method presented in this paper is provided as in Table \ref{result}.

\begin{table*}[htb]
\renewcommand{\arraystretch}{1.3}
\setlength{\tabcolsep}{6pt}
\centering

\begin{tabular}{c c c c c c c c c}
\hline
Round & Alternative  & & x1 & x2 & x3 & x4 &x5 &x6 \\
\hline
r=1 & $Supplier_1$ & $Expert_1$ & ((0.6,0.2),0.6400) & ((0.3,0.5),0.3250) & ((0.7,0.1),0.7650) & ((0.6,0.1),0.7200) &((0.5,0.3),0.5250) &-\\
& &$Expert_2$ & ((0.3,0.4),0.3900) & ((0.4,0.4),0.4200) & ((0.5,0.2),0.6000) &( (0.7,0.2),0.6800) & ((0.5,0.3),0.5250) &-\\
& &$Expert_3$ & ((0.5,0.4),0.4500) & ((0.4,0.3),0.4900) & ((0.6,0.1),0.7200) &( (0.5,0.3),0.5250) & ((0.3,0.4),0.3900) &-\\

& $Supplier_2$ & $Expert_1$ & ((0.8,0.0),0.9000) & ((0.3,0.6),0.2600) & ((0.5,0.4),0.4500) & ((0.3,0.5),0.3250) &((0.6,0.2),0.6400) &-\\
& &$Expert_2$ & ((0.7,0.3),0.5950) & ((0.4,0.5),0.3500) & ((0.4,0.3),0.4900) & ((0.6,0.1),0.7200) & ((0.5,0.2),0.6000) &-\\
& &$Expert_3$ & ((0.8,0.1),0.8100) & ((0.6,0.2),0.6400) & ((0.4,0.5),0.3500) & ((0.4,0.4),0.4200) & ((0.6,0.1),0.7200) &-\\

& $Supplier_3$ & $Expert_1$ & ((0.3,0.6),0.2600) & ((0.7,0.0),0.8500) & ((0.6,0.2),0.6400) & ((0.5,0.3),0.5250) &((0.6,0.2),0.6400) &-\\
& &$Expert_2$ & ((0.5,0.2),0.6000) & ((0.6,0.3),0.5600) & ((0.5,0.1),0.6750) & ((0.4,0.4),0.4200) & ((0.8,0.2),0.7200) &-\\
& &$Expert_3$ & ((0.7,0.2),0.6800) & ((0.5,0.2),0.6000) & ((0.6,0.3),0.5600) & ((0.9,0.1),0.8550) & ((0.7,0.1),0.7650) &-\\

& $Supplier_4$ & $Expert_1$ & ((0.3,0.6),0.2600) & ((0.7,0.2),0.6800) &((0.6,0.2),0.6400) & ((0.7,0.1),0.7650) &((0.7,0.1),0.7650) &-\\
& &$Expert_2$ & ((0.7,0.1),0.7650) &((0.7,0.2),0.6800) & ((0.5,0.2),0.6000) & ((0.7,0.1),0.7650) & ((0.7,0.1),0.7650) &-\\
& &$Expert_3$ & ((0.7,0.1),0.7650) & ((0.7,0.2),0.6800) & ((0.8,0.1),0.8100) & ((0.7,0.1),0.7650) & ((0.7,0.2),0.6800) &-\\

& $Supplier_5$ & $Expert_1$ & ((0.4,0.5),0.3500) & ((0.6,0.2),0.6400)& ((0.3,0.5),0.3250) & ((0.3,0.5),0.3250) &((0.4,0.3),0.4900) &-\\
& &$Expert_2$ & ((0.4,0.4),0.4200) & ((0.5,0.3),0.5250) & ((0.2,0.3),0.4200) & ((0.5,0.3),0.5250) & ((0.6,0.2),0.6400) &-\\
& &$Expert_3$ & ((0.6,0.2),0.6400) & ((0.5,0.3),0.5250) & ((0.6,0.2),0.6400) & ((0.6,0.2),0.6400) & ((0.5,0.3),0.5250) &-\\
\hline
\end{tabular}
\caption{ Transformed information about the judgments from experts in the first round}
\label{RA0.6}
\end{table*}

\begin{table*}[htb]
\renewcommand{\arraystretch}{1.3}
\setlength{\tabcolsep}{7pt}
\centering
\begin{tabular}{c c c c c c c c c}
\hline
Round & Alternative  & & x1 & x2 & x3 & x4 &x5 &x6 \\
\hline
r=1 & $Supplier_1$ & $Expert_1$ & (0.3840,0.1280) & (0.0975,0.1625) & (0.5355,0.0765) & (0.4320,0.0720) &(0.2625,0.1575) &-\\
& &$Expert_2$ & (0.1170,0.1560) & (0.1680,0.1680) & (0.3000,0.1200) &(0.4760,0.1360) & (0.2625,0.1575) &-\\
& &$Expert_3$ & (0.2250,0.1800) & (0.1960,0.1470) & (0.4320,0.0720) &(0.2625,0.1575) & (0.1170,0.1560) &-\\

& $Supplier_2$ & $Expert_1$ & (0.7200,0.0000) & (0.0780,0.1560) & (0.2250,0.1800) & (0.0975,0.1625) &(0.3840,0.1280) &-\\
& &$Expert_2$ & (0.4165,0.1785) & (0.1400,0.1750) & (0.1960,0.1470) &(0.4320,0.0720) & (0.3000,0.1200) &-\\
& &$Expert_3$ & (0.6480,0.0810) & (0.3840,0.1280) & (0.1400,0.1750) &(0.1680,0.1680) & (0.4320,0.0720) &-\\

& $Supplier_3$ & $Expert_1$ & (0.0780,0.1560) & (0.5950,0.0000) & (0.3840,0.1280) & (0.2625,0.1575) &(0.3840,0.1280) &-\\
& &$Expert_2$ & (0.3000,0.1200) & (0.3360,0.1680) & (0.3375,0.0675) &(0.1680,0.1680) & (0.5760,0.1440) &-\\
& &$Expert_3$ & (0.5355,0.0765) & (0.4760,0.1360) & (0.3000,0.1200) &(0.5355,0.0765) & (0.5355,0.0765) &-\\

& $Supplier_4$ & $Expert_1$ & (0.3360,0.1680) & (0.4760,0.1360) & (0.3840,0.1280) & (0.5355,0.0765) &(0.5355,0.0765) &-\\
& &$Expert_2$ & (0.5355,0.0765) & (0.4760,0.1360) & (0.3000,0.1200) &(0.5355,0.0765) & (0.5355,0.0765) &-\\
& &$Expert_3$ & (0.5355,0.0765) & (0.4760,0.1360) & (0.6480,0.0810) &(0.5355,0.0765) & (0.5355,0.0765) &-\\

& $Supplier_5$ & $Expert_1$ & (0.1400,0.1750) & (0.3840,0.1280) & (0.0975,0.1625) & (0.0975,0.1625) &(0.1960,0.1470) &-\\
& &$Expert_2$ & (0.1680,0.1680) & (0.2625,0.1575) & (0.0840,0.1260) &(0.2625,0.1575) & (0.3840,0.1280) &-\\
& &$Expert_3$ & (0.3840,0.1280) & (0.2625,0.1575) & (0.3840,0.1280) &(0.3840,0.1280) & (0.2625,0.1575) &-\\
\hline
\end{tabular}
\caption{Transformed information about the judgments from experts in the first round}
\label{RA0.38}
\end{table*}

\subsection{Round 1}
In the first round, detailed process of generating final judgments on situation of suppliers is provided.

\textbf{Step 1}: Obtain information in the form of IFS.

\textbf{Step 2}: For example, in regard to $B = (Expert_1, x1)$, the corresponding value of score function can be obtained by
$SF(B)=\frac{1}{2}(1+0.6)(1-0.2)=0.64$. Then, the values of IFSs are provided in Table\ref{tab0.64}.
\begin{table}
\renewcommand{\arraystretch}{1.5}
\setlength{\tabcolsep}{6pt}
\centering

\begin{tabular}{c c c c c c c }
\hline
Alternative  & & x1 & x2 & x3 & x4 &x5  \\
\hline
$Supplier_1$ &  $Expert_1$ & 0.64 &0.325 &0.765  &0.72  &0.525 \\

 &Expert2 & 0.39 & 0.42 & 0.6 & 0.68 & 0.525   \\

 &Expert3 &0.45  & 0.49 & 0.72 & 0.525 &0.39 \\

 $Supplier_1$ & $Expert_1$ & 0.9& 0.26& 0.45 & 0.325 &0.64 \\

 &Expert2 &0.595& 0.35 & 0.49 & 0.72 & 0.6   \\

 &Expert3 & 0.81 & 0.64 & 0.35 & 0.42 & 0.72\\

 $Supplier_1$ & $Expert_1$ &0.26 &0.85 &0.64  &0.525  &0.64 \\

 &Expert2 & 0.6 &0.56  &0.675  &0.42 &0.72  \\

 &Expert3 &0.68  &0.6  &0.56  &0.855  &0.765 \\

 $Supplier_1$ & $Expert_1$ &0.56 &0.68 & 0.64 & 0.765 & 0.765\\

 &Expert2 & 0.765 & 0.68 & 0.6 & 0.765 &0.765    \\

 &Expert3 &0.765  &0.68  &0.81  &0.765  &0.68 \\

 $Supplier_1$ & $Expert_1$ &0.35 &0.64 &0.325  &0.325  &0.49 \\

 &Expert2 & 0.42 &0.525  &0.42  &0.525  &0.64    \\

 &Expert3 &0.64 & 0.525 & 0.64 &0.64  &0.525 \\
 
\hline
\end{tabular}
\caption{Values of score function of IFSs}
\label{tab0.64}
\end{table}

The motivation to produce reliability of each IFS is for reduce the influence brought by unreliable ones, which is devised to avoid producing biased results.

\textbf{Step 3}: Using the values obtained in the last step, Z numbers can be constructed. The results are given in Table \ref{RA0.6}.

\textbf{Step 4}: Combine the transformed numbers. And the results are provided in Table \ref{RA0.38}.

\textbf{Step 5}: Calculate distances of IFSs within groups. Besides, the results
are given in Table \ref{RA0}.

This operation is carried out to alleviate the negative effects led by the isolated individuals of IFS which do not conform to the trend exerted by the main body of information.

\begin{center}
\setlength{\tabcolsep}{6pt}
\renewcommand{\arraystretch}{1.40}
{\footnotesize
\topcaption{Distances of IFSs within groups (Round1)}
\label{RA0}
\begin{supertabular}{c c c c c c c}
\\

\hline
 Alternative  & & x1 & x2 & x3 & x4 &x5 \\
\hline

\multirow{5}{*}{$Supplier_1$} 
&x1 & 0 & 0.2442 & 0.1114 &0.0685 & 0.0921\\
 & x2 & 0.2442 &  0& 0.3453 & 0.2795& 0.1566\\
 & x3 & 0.1114 & 0.3453 & 0.2795  &0.1566 & \\ 
 & x4 & 0.0685 & 0.2795 & 0.0782 & 0 &0.1423 \\
 & x5 & 0.0921 & 0.1566 & 0.2014 & 0.1423 & 0 \\
\hline

 \multirow{5}{*}{$Supplier_1$} 
& x1 & 0 & 0.0558 & 0.1615 &0.2912 &0.1358 \\
& x2 &0.0558  &0  &0.1134  &0.2399 & 0.0821\\
 & x3 &0.1615  &0.1134  &0  &0.1411 &0.1601 \\ 
 & x4 & 0.0338 & 0.0609 & 0.1423 &0 & 0.1358 \\
 & x5 & 0.1114 & 0.0774 & 0.2588 & 0.1358 &0 \\
\hline

 \multirow{5}{*}{$Supplier_1$} 
&  x1 &0  &0.0458  &0.1755  &0.0338 &0.1114 \\
 & x2 &0.0458  &0  &0.1869  &0.0609 &0.0774 \\
 & x3 &0.1755  &0.1869  &0  &0.1423 &0.2588 \\ 
 & x4 &0.494  &0.0262  &0.1308  &0 &0.2442 \\
 & x5 &0.2909  &0.2673  &0.1239  &0.2442 &0 \\
\hline

\multirow{5}{*}{$Supplier_2$} 
& x1 & 0  &0.5115  &0.3994  &0.494 &0.2909 \\
 & x2 & 0.2326  &0  &0.0557  &0.2403 &0.1402 \\
 & x3 &0.1901  &0.0557  &0  &0.2403 &0.0859 \\ 
 & x4 &0.1186  &0.2403  &0.1869  &0 &0.1032 \\
 & x5 &0.1243  &0.1402  &0.0859  &0.1032 &0 \\
\hline

\multirow{5}{*}{$Supplier_2$} 
& x1 &   0  &0.2326  &0.1901  &0.1186 &0.1243 \\
 & x2 & 0.2326  &0  & 0.0557 &0.2403 &0.1402 \\
 & x3 &0.1901  &0.0557  &0  &0.1869 &0.0859 \\ 
 & x4 &0.1186  &0.2403  &0.1869  &0 &0.1032 \\
 & x5 &0.1243 &0.1402  &0.0859  &0.1032 &0 \\ 
\hline

\multirow{5}{*}{$Supplier_2$} 
& x1 &0  &0.1883  &0.3794  &0.3546 &0.1666 \\
&  x2 &0.1883  &0  & 0.1992 &0.1727&0.0685 \\
& x3 &0.3794  &0.1992  &0  &0.0275 &0.2403 \\ 
 & x4 &0.3546  &0.1727  &0.0275  &0 &0.2146 \\
 & x5 &1666  &0.0685  &0.2403  &0.2146 &0 \\
\hline

\multirow{5}{*}{$Supplier_3$} 
&  x1 & 0& 0.4415 &0.2673  &0.1812 &0.2673 \\
 & x2 & 0.4415 & 0 & 0.2403  &0.309 &0.2403 \\
 & x3 &0.2673  &0.2403  &0  &0.0921 &0 \\ 
 & x4 &0.1812  &0.309  & 0.0921 &0 &0.0921 \\
 & x5 & 0.2673 &0.2403  &0  &0.0921 &0 \\
\hline

\multirow{5}{*}{$Supplier_3$} 
&x1 & 0 & 0.0645 &0.0658  &0.1134 &0.2222 \\
& x2 & 0.0645 &0  &0.1154  &0.1427 &0.178 \\
 & x3 &0.0658  &0.1154  &0  &0.1628 &0.2276 \\ 
& x4 &0.1134  &0.1427  &0.1628  &0 &0.3159 \\
 & x5 &0.2222  &0.178  &0.2276  &0.3159 &0 \\
\hline

\multirow{5}{*}{$Supplier_3$} 
&  x1 & 0 &0.1411  &0.1011  &0.2216 &0.0714 \\
 & x2 & 0.1411  & 0 &0.0645  &0.3531 &0.1698 \\
 & x3 &0.1011  &0.0645  &0  &0.3183 &0.1536 \\ 
 & x4 &0.2216  &0.3531  &0.3183  &0 &0.1985 \\
 & x5 &0.0714  &0.1698  &0.1536  &0.1985 &0 \\
\hline

\multirow{5}{*}{$Supplier_4$} 
&x1 & 0 & 0.1011 & 0.0466  &0.1536 & 0.1536\\
 & x2 & 0.1011 & 0 & 0.0734 & 0.0714 &0.0714 \\
 & x3& 0.0466 & 0.0734 & 0 & 0.1114&0.1114 \\ 
 & x4 & 0.1536 & 0.0714 & 0.1114 & 0 &0 \\
 & x5& 0.1536 & 0.0714 & 0.1114 & 0 & 0\\
\hline

\multirow{5}{*}{$Supplier_4$} 
& x1 & 0 & 0.0714 & 0.1698 & 0 & 0\\
 & x2& 0.0714 & 0 & 0.1411 & 0.0714&0.0714 \\
 & x3 & 0.1698 & 0.1411 & 0 & 0.1698&.1698 \\ 
 & x4 & 0 & 0.0714 & 0.1698 & 0 &0 \\
 & x5 & 0 & 0.0714 & 0.1698 & 0&0 \\
\hline

\multirow{5}{*}{$Supplier_4$} 
& x1 & 0 & 0.0714 &0.0891  &0 &0.0714 \\
 & x2 &0.0714  &0  &0.1239  &0.0714 &0 \\
 & x3 &0.0891  &0.1239  &0  &0.0891 &0.1239 \\ 
 & x4 &0  &0.0714  &0.0891  &0 &0.0714 \\
 & x5 &0.0714  &0  &0.1239  &0.0714 &0 \\
\hline

\multirow{5}{*}{$Supplier_5$} 
&  x1 & 0 & 0.1992 & 0.0506 & 0.0506 & 0.0557\\
 & x2 & 0.1992 & 0 & 0.2442 &0.2442 &  0.1487\\
 & x3& 0.0506 & 0.2442 & 0 & 0 & 0.0993 \\ 
 & x4 & 0.0506 & 0.2442 & 0 & 0 & 0.0993 \\
 & x5 & 0.0557 & 0.1487 & 0.0993 & 0.0993&0 \\
\hline

\multirow{5}{*}{$Supplier_5$} 
&  x1 & 0 & 0.0821 & 0.1066 & 0.0821 &0.1727\\
 & x2& 0.0821 & 0 & 0.1819  &0 & 0.0921  \\
 & x3 & 0.1066 & 0.1819 & 0 & 0.1819 & 0.2647\\ 
 & x4 & 0.0821 & 0 & 0.1819 &0 &0.0921\\
 & x5 & 0.1727 & 0.0921 & 0.2647 & 0.0921 & 0\\
\hline

\multirow{5}{*}{$Supplier_5$} 
& x1 & 0 & 0.0921 & 0 &0 &0.0921\\
 & x2& 0.0921 & 0 & 0.0921 & 0.0921&0 \\
 & x3 & 0 & 0.0921 & 0 & 0.0921 & 0\\ 
 & x4 & 0 & 0.0921 & 0 &0 & 0.0921\\
 & x5& 0.0921 & 0 & 0.0921 & 0.0921&0  \\
\hline
\end{supertabular}
}
\end{center}

\textbf{Step 6}: Utilize the closeness centrality to obtain similarity measure of
IFSs within groups. And the results are offered in Table \ref{A7.7}.

The introduction of closeness centrality indirectly to construct a series of simple network for groups of IFSs, which provide a convenient method in measuring degree of importance of nodes which is consisted of membership and non-membership of IFSs.

\vspace{-0.5cm}
\begin{center}
\begin{table}[H]
\renewcommand{\arraystretch}{1.3}
\setlength{\tabcolsep}{6pt}
\centering
\begin{tabular}{c c c c c c}
\hline
Alternative &x1 &x2&x3 &x4 &x5\\
\hline
\multirow{3}{*}{$Supplier_1$}& 7.7485 & 3.9002 & 5.4325 & 7.0353 & 6.7521\\
& 6.2072 & 8.1436 & 8.7036 & 4.8064 & 9.4891\\
&10.9134&10.7812&5.2398&10.7274&6.8566\\
\hline
\multirow{3}{*}{$Supplier_2$}&2.3588 & 4.1611 & 4.9355 & 4.4679 & 4.3184 \\
& 6.0091 & 5.9812 & 7.7137 & 6.1635 & 8.8186 \\
&3.6734 &6.3625& 4.7261& 5.1998&5.7973\\
\hline
\multirow{3}{*}{$Supplier_3$} & 8.7929 & 12.6087 & 11.6697 & 11.8908 & 11.8908 \\
& 16.584 & 11.2598 & 6.1493 & 16.584 & 16.584 \\
&7.4749&5.4913&6.275&3.6648&6.7414\\
\hline
\multirow{3}{*}{$Supplier_4$}& 8.7929 & 12.6087 & 11.6697 & 11.8908 & 11.8908 \\
& 16.584 & 11.2598 & 6.1493 & 16.584 & 16.584 \\
& 17.2536 & 14.9994 & 9.392 & 17.2536 & 14.9994 \\
\hline
\multirow{3}{*}{$Supplier_5$}& 11.2355 & 4.7835 & 10.1516 & 10.1516 & 9.9286 \\ 
& 9.0211 & 11.2331 & 5.4408 & 11.2331 & 6.4342 \\ 
& 21.7061 & 14.4708 & 21.7061 & 21.7061 & 14.4708 \\ 
\hline
\end{tabular}
\caption{ Degree of similarity measure of IFSs within groups}
\label{A7.7}
\end{table}
\end{center}
\vspace{-0.5cm}
\vspace{-0.7cm}

\textbf{Step 7}: Then, generate the fuzzy relationship based on the similarity
measure. The corresponding fuzzy relationship matrices can be given as follows:

\begin{equation}
\scriptsize{
\setlength{\arraycolsep}{2.5pt}
a_1^{E_1}=\begin{bmatrix}0&1&1&1&1\\0&0&0&0&0\\0&1&0&0&0\\0&1&1&0&1\\0&1&1&0&0\end{bmatrix}a_1^{E_2}=\begin{bmatrix}0&0&0&1&0\\1&0&0&1&0\\1&1&0&1&0\\0&0&0&0&0\\1&1&1&1&0\end{bmatrix}a_1^{E_3}=\begin{bmatrix}0&1&1&1&1\\0&0&1&1&1\\0&0&0&0&0\\0&0&1&0&1\\0&0&1&0&0\end{bmatrix}\notag
}
\end{equation}

\begin{equation}
\scriptsize{
\setlength{\arraycolsep}{2.5pt}
a_2^{E_1}=\begin{bmatrix}0&0&0&0&0\\1&0&0&0&0\\1&1&0&1&1\\1&1&0&1&1\\1&1&0&0&1\\1&1&0&0&0\end{bmatrix}a_2^{E_2}=\begin{bmatrix}0&1&0&0&0\\0&0&0&0&0\\1&1&0&1&0\\1&1&0&0&0\\1&1&0&0&0\\1&1&1&1&0\end{bmatrix}a_2^{E_3}=\begin{bmatrix}0&0&0&0&0\\1&0&1&1&1\\1&0&0&0&0\\1&0&1&0&0\\1&0&1&1&0\end{bmatrix}\notag
}
\end{equation}

\begin{equation}
\scriptsize{
\setlength{\arraycolsep}{2.5pt}
a_3^{E_1}=\begin{bmatrix}0&1&0&0&0\\0&0&0&0&0\\1&1&0&1&0\\1&1&0&0&0\\1&1&0&0&0\\1&1&0&1&0
\end{bmatrix}
a_3^{E_2}=\begin{bmatrix}0&1&1&1&1\\0&0&1&1&1\\0&0&1&1&1\\0&0&0&1&1\\
0&0&0&0&1\\0&0&0&0&0
\end{bmatrix}
a_3^{E_3}=\begin{bmatrix}0&1&1&1&1\\0&0&0&1&0\\0&0&0&1&0\\0&1&0&1&0\\0&0&0&0&0\\
0&1&1&1&0\end{bmatrix}\notag
}
\end{equation}

\begin{equation}
\scriptsize{
\setlength{\arraycolsep}{2.5pt}
a_4^{F_1}=\begin{bmatrix}0&0&0&0&0\\1&0&1&1&1\\1&0&0&0&0\\1&0&1&0&0\\1&0&1&0&0\\1&0&1&0&0\end{bmatrix}a_4^{E_2}=\begin{bmatrix}0&1&1&0&0\\0&0&1&0&0\\0&0&0&0&0\\0&1&1&0&0\\0&1&1&0&0\end{bmatrix}a_4^{E_3}=\begin{bmatrix}0&1&1&0&1\\0&0&1&0&0\\0&0&0&0&0\\0&1&1&0&1\\0&0&1&0&0\end{bmatrix}\notag
}
\end{equation}

\begin{equation}
\scriptsize{
\setlength{\arraycolsep}{2.5pt}
a_5^{E_1}=\begin{bmatrix}0&1&1&1&1\\0&0&0&0&0\\0&1&0&0&1\\0&1&0&0&1\\0&1&0&0&1\\0&1&0&0&0\end{bmatrix}a_5^{E_2}=\begin{bmatrix}0&0&1&0&1\\1&0&1&0&1\\0&0&0&0&0\\1&0&1&0&1\\0&0&1&0&0\end{bmatrix}a_5^{E_3}=\begin{bmatrix}0&1&0&0&1\\0&0&0&0&0\\0&1&0&0&1\\0&1&0&0&1\\0&1&0&0&1\\0&0&0&0&0\end{bmatrix}\notag
}
\end{equation}

And the points obtained by each IFS is given in Table \ref{A4}.

The 0-1 relationship is introduced based on divergence of individual IFSs within one group to serve as a selective tool in distinguishing important information and erasing useless part.

\textbf{Step 8}: Compute the weights which is based on the points obtained in \textbf{Step 7}. The results of the weights are given in Table \ref{A0.4}.

\textbf{Step 9}: Calculate the distances between groups of IFSs. The corresponding values of distances are given in Table \ref{ADD}.

\textbf{Step 10 and 11}: Compute the degree of credibility and normalize them. And the values of degree of credibility is provided in Table \ref{A0.32}.

After evaluating difference of individuals and groups, the final degree of credibility can be obtained.

\begin{center}

\begin{table}[htb]
\renewcommand{\arraystretch}{1.2}
\setlength{\tabcolsep}{14pt}
\centering
\begin{tabular}{c c c c c c}
\hline
Alternative &x1 &x2&x3 &x4 &x5\\
\hline
$Supplier_1$ &4 &0 & 1 &3 &2\\
&1 &2 &3 &0 &4\\
&4&3&0&2&l\\
$Supplier_2$&0&1&4&3&2\\
&1&0&3&2&4\\&0&4&1&2&3\\
$Supplier_3$&l&0&3&2&3\\
&4&3&2&1&0\\
&4&1&2&0&3\\
$Supplier_4$&0&4&1&2&2\\
&2&1&0&2&2\\
&3&1&0&3&1\\
$Supplier_5$&4&0&2&2&1\\
&2&3&0&3&1\\
&2&0&2&2&0\\
\hline
\end{tabular}
\caption{Points obtained by each IFS}
\label{A4}
\end{table}
\end{center}

\begin{center}
\begin{table}[htb]
\renewcommand{\arraystretch}{1.3}
\setlength{\tabcolsep}{10pt}
\centering
\begin{tabular}{c c c c c c}
\hline
Alternative &x1 &x2&x3 &x4 &x5\\
\hline
$Supplier_1$& 0.4000 & 0.0000 & 0.1000 & 0.3000 & 0.2000 \\
& 0.1000 & 0.2000 & 0.3000 & 0.0000 & 0.4000 \\
&0.4000&0.3000&0.0000&0.2000&0.1000\\

$Supplier_2$& 0.0000 & 0.1000 & 0.4000 & 0.3000 & 0.2000 \\ 
& 0.1000 & 0.0000 & 0.3000 & 0.2000 & 0.4000 \\ 
&0.0000 &0.4000&0.1000&0.2000&0.3000\\

$Supplier_3$& 0.1111 & 0.0000 & 0.3333 & 0.2222 & 0.3333 \\
& 0.4000 & 0.3000 & 0.2000 & 0.1000 & 0.0000 \\
&0.4000 &0.1000&0.2000&0.0000&0.3000\\

$Supplier_4$& 0.0000 & 0.4444 & 0.1111 & 0.2222 & 0.2222 \\ 
& 0.2857 & 0.1429 & 0.0000 & 0.2857 & 0.2857 \\
& 0.3750 & 0.1250 & 0.0000 & 0.3750 & 0.1250 \\

$Supplier_5$& 0.4444 & 0.0000 & 0.2222 & 0.2222 & 0.1111  \\ 
& 0.2222 & 0.3333 & 0.0000 & 0.3333 & 0.1111 \\ 
& 0.3333 & 0.0000 & 0.3333 & 0.3333 & 0.0000 \\ 
\hline
\end{tabular}
\caption{Weight obtained by points}
\label{A0.4}
\end{table}
\end{center}
\vspace{-0.5cm}
\vspace{-0.5cm}
\vspace{-0.5cm}

\textbf{Step 12}: Calculate the information volume of each group of IFSs and the values of information volumes are offered in Table \ref{A8.5}.

\begin{center}
\begin{table}[htb]
\renewcommand{\arraystretch}{1.33}
\setlength{\tabcolsep}{5pt}
\centering
\begin{tabular}{c c c c c c c}
\hline
Alternative &Distances &x1 &x2&x3 &x4 &x5\\
\hline

$Supplier_1$& $D_{E_{1-2}}$&0.4000 & 0.0000 & 0.1000 & 0.3000 & 0.2000 \\
& $D_{E_{1-3}}$&0.0496 & 0.0000 & 0.0078 & 0.0427 & 0.0272 \\

$Supplier_1$ & $D_{E_{2-1}}$&0.0224&0.0152&0.0509&0.0000&0.0000\\
& $D_{E_{2-3}}$& 0.0111&0.0060&0.0310&0.0000&0.0543 \\

$Supplier_1$& $D_{E_{3-1}}$ & 0.0496 & 0.0298 & 0.0000 & 0.0285 & 0.0136 \\
& $D_{E_{3-2}}$& 0.0445 & 0.0090 & 0.0000 & 0.0320 & 0.0136 \\

$Supplier_2$& $D_{E_{1-2}}$&0.0000 & 0.0267 & 0.0325 & 0.0228 & 0.0137 \\
& $D_{E_{1-3}}$& 0.0000 & 0.0267 & 0.0325 & 0.0228 & 0.0137 \\

$Supplier_2$ & $D_{E_{2-1}}$& 0.0297 & 0.0000 & 0.0138 & 0.0429 & 0.0413 \\ 
& $D_{E_{2-3}}$& 0.0170 & 0.0000 & 0.0167 & 0.0429 & 0.0413 \\

$Supplier_2$& $D_{E_{3-1}}$ & 0.0000 & 0.1069 & 0.0081 & 0.0152 & 0.0206 \\
& $D_{E_{3-2}}$& 0.0000 & 0.0797 & 0.0056 & 0.0429 & 0.0310 \\

$Supplier_3$& $D_{E_{1-2}}$&0.0229 & 0.0000 & 0.0301 & 0.0128 & 0.0518 \\
& $D_{E_{1-3}}$& 0.0371 & 0.0000 & 0.0301 & 0.0182 & 0.0518 \\

$Supplier_3$ & $D_{E_{2-1}}$& 0.0824 & 0.0842 & 0.0181 & 0.0082 & 0.0000 \\ 
& $D_{E_{2-3}}$& 0.0564 & 0.0193 & 0.0231 & 0.0449 & 0.0000 \\

$Supplier_3$& $D_{E_{3-1}}$ & 0.1334 & 0.0272 & 0.0093 & 0.0000 & 0.0334 \\
& $D_{E_{3-2}}$& 0.0564 & 0.0064 & 0.0231 & 0.0000 & 0.0300 \\

$Supplier_4$& $D_{E_{1-2}}$&0.0000 & 0.0000 & 0.0076 & 0.0000 & 0.0000 \\
& $D_{E_{1-3}}$& 0.0000 & 0.0000 & 0.0209 & 0.0000 & 0.0000 \\

$Supplier_4$ & $D_{E_{2-1}}$& 0.0439 & 0.0000 & 0.0000& 0.0000 & 0.0000 \\ 
& $D_{E_{2-3}}$& 0.0000 & 0.0000 & 0.0000 & 0.0000 & 0.0204 \\

$Supplier_4$& $D_{E_{3-1}}$ & 0.0576 & 0.0000 & 0.0000 & 0.0000 & 0.0089 \\
& $D_{E_{3-2}}$& 0.0000 & 0.0000 & 0.0000 & 0.0000 & 0.0089  \\

$Supplier_5$& $D_{E_{1-2}}$&0.0122 & 0.0000 & 0.0095 & 0.0348 & 0.0165 \\
& $D_{E_{1-3}}$& 0.0885 & 0.0000 & 0.0543 & 0.0543 & 0.0068 \\

$Supplier_5$ & $D_{E_{2-1}}$& 0.0061 & 0.0307 & 0.0000& 0.0522 & 0.0165 \\ 
& $D_{E_{2-3}}$& 0.0384 & 0.0000 & 0.0000 & 0.0307 & 0.0102 \\

$Supplier_5$& $D_{E_{3-1}}$ & 0.0664 & 0.0000 & 0.0814 & 0.0814 & 0.0000 \\
& $D_{E_{3-2}}$& 0.0575 & 0.0000 & 0.0882 & 0.0307 & 0.0000  \\
\hline

\end{tabular}
\caption{Weight obtained by points}
\label{ADD}
\end{table}
\end{center}

\begin{table}[ht]
\renewcommand{\arraystretch}{1.2}
\setlength{\tabcolsep}{14pt}
\centering

\begin{tabular}{c c c c}
\hline
Alternative  &$Expert_1$ &$Expert_2$&$Expert_3$ \\
\hline
$Supplier_1$ &0.3221 &0.3434 &0.3345\\
$Supplier_2$ &0.3453 &0.3375 &0.3172 \\
$Supplier_3$ &0.3403 &0.3270 &0.3327 \\
$Supplier_4$ &0.3378 &0.3326 &0.3296 \\
$Supplier_5$ &0.3373 &0.3666 &0.2962 \\
\hline
\end{tabular}
\caption{Degree of credibility of groups of IFSs}
\label{A0.32}
\end{table}

\begin{table}[htb]
\renewcommand{\arraystretch}{1.2}
\setlength{\tabcolsep}{14pt}
\centering

\begin{tabular}{c c c c}
\hline
Alternative  &$Expert_1$ &$Expert_2$&$Expert_3$ \\
\hline
$Supplier_1$ &8.5376 &8.9641 &9.1858\\
$Supplier_2$ &7.5027 &8.1791 &7.8977 \\
$Supplier_3$ &7.9891 &7.9707 &6.6730 \\
$Supplier_4$ &7.4047 &7.6976 &6.6585 \\
$Supplier_5$ &8.8588 &9.4098 &8.6688 \\
\hline
\end{tabular}
\caption{Information volume of each group of IFSs}
\label{A8.5}

\end{table}

\textbf{Step 13 and 14}: Modify and normalize the information volume. The results are provided in Table \ref{A0.22}.
\begin{table}[htb]
\renewcommand{\arraystretch}{1.2}
\setlength{\tabcolsep}{14pt}
\centering

\begin{tabular}{c c c c}
\hline
Alternative  &$Expert_1$ &$Expert_2$&$Expert_3$ \\
\hline
$Supplier_1$ &0.2250 &0.3447 &0.4303\\
$Supplier_2$ &0.2247 &0.4419 &0.3335 \\
$Supplier_3$ &0.4445 &0.4364 &0.1192 \\
$Supplier_4$ &0.3553 &0.4762 &0.1685 \\
$Supplier_5$ &0.2807 &0.4871 &0.2322 \\
\hline
\end{tabular}
\caption{Processed information volume of each group of IFSs}
\label{A0.22}
\end{table}

Information volume is an index of the degree of uncertainty of IFS. This step utilize the concept of the entropy method to further optimize the effect of strategy designing.

\textbf{Step 15 and 16}: Generate the attitude character and normalize them.

The modified attitude characters are provided in Table \ref{A0.21}.
\begin{table}[htb]
\renewcommand{\arraystretch}{1.2}
\setlength{\tabcolsep}{14pt}
\centering

\begin{tabular}{c c c c}
\hline
Alternative  &$Expert_1$ &$Expert_2$&$Expert_3$ \\
\hline
$Supplier_1$ &0.2165 &0.3536 &0.4299\\
$Supplier_2$ &0.2333 &0.4485 &0.3181 \\
$Supplier_3$ &0.4534 &0.4277 &0.1189 \\
$Supplier_4$ &0.3594 &0.4743 &0.1663 \\
$Supplier_5$ &0.2768 &0.5221 &0.2011 \\
\hline
\end{tabular}
\caption{Attitude characters of each group of IFSs}
\label{A0.21}
\end{table}

\textbf{Step 17}: According to the attitude characters obtained, the dynamic formula for determining the expression of OWA operator utilizing varied parameters can be constructed as follows:
\begin{equation}
\begin{gathered}
Supplier_1 :(\frac sk)^{\frac{1-\alpha_{1}^{Nor}}{\alpha_{1}^{Nor}}}-(\frac{s-1}k)^{\frac{1-\alpha_{1}^{Nor}}{\alpha_{1}^{Nor}}} \\
Supplier_2 :(\frac sk)^{\frac{1-\alpha_{2}^{Nor}}{\alpha_{2}^{Nor}}}-(\frac{s-1}k)^{\frac{1-\alpha_{2}^{Nor}}{\alpha_{2}^{Nor}}} \\
Supplier_3 :(\frac sk)^{\frac{1-\alpha_{3}^{Nor}}{\alpha_{3}^{Nor}}}-(\frac{s-1}k)^{\frac{1-\alpha_{3}^{Nor}}{\alpha_{3}^{Nor}}} \\
Supplier_4 :(\frac sk)^{\frac{1-\alpha_{4}^{Nor}}{\alpha_{4}^{Nor}}}-(\frac{s-1}k)^{\frac{1-\alpha_{4}^{Nor}}{\alpha_{4}^{Nor}}} \\
Supplier_5 :(\frac Sk)^{\frac{1-\alpha_{5}^{Nor}}{\alpha_{5}^{Nor}}}-(\frac{s-1}k)^{\frac{1-\alpha_{5}^{Nor}}{\alpha_{5}^{Nor}}} 
\end{gathered} \notag
\end{equation}

\textbf{Step 18}: Calculate the corresponding weights using improved method of generating OWA operators which are provided in Table \ref{AE3.61}.

\begin{table}[htb]
\scriptsize{
\renewcommand{\arraystretch}{1.4}
\setlength{\tabcolsep}{5pt}
\centering
\begin{tabular}{c c c c}
\hline
Alternative  &$Expert_1$ &$Expert_2$&$Expert_3$ \\
\hline
$Supplier_1$ & 3.618937644341801 & 1.8280542986425337 & 1.3261223540358225 \\
$Supplier_2$ &3.2863266180882977 &1.229654403567447 &2.1436655139893115 \\
$Supplier_3$ &1.2055580061755622 &1.3380874444704232 &7.410428931875526 \\
$Supplier_4$ &1.7824151363383418 &1.1083702298123552 &5.013229104028864 \\
$Supplier_5$ &2.6127167630057806 &0.9153418885271021 &3.9726504226752857 \\
\hline
\end{tabular}
}
\caption{Corresponding weights of groups of IFSs}
\label{AE3.61}
\end{table}

\textbf{Step 19}: Allocate a mark number to each of IFS based on its sequence which is within groups of judgments from experts.

\textbf{Step 20, 21, 22, 23}: Based on the expressions of soft likelihood function, the weights for each IFS within groups can be obtained. And then, divide the degree of hesitancy into membership and non-membership and input them into the improved soft likelihood function to generate the final judgments for gross estimations from experts. The gross estimation with respect
to suppliers are given in Table \ref{AD5} and Figure \ref{fig_2}.

\begin{table}[htb]
\renewcommand{\arraystretch}{1.3}
\setlength{\tabcolsep}{13pt}
\centering
\begin{tabular}{c c c c c}
\hline
&Alternative  &Degree of preference \\
\hline
&$Supplier_1$ &5.659509966930474 \\
&$Supplier_2$ &8.792310830908539 \\
&$Supplier_3$ &11.779741914481567 \\
&$Supplier_4$ &16.267048625369885 \\
&$Supplier_5$ &4.6228307296199755 \\
\hline
\multicolumn{3}{c}{$Supplier_4>Supplier_3>Supplier_2>Supplier_1>Supplier_5$}\\
\hline
\end{tabular}
\caption{Gross estimation with respect to suppliers}
\label{AD5}
\end{table}

\begin{figure}[htb]
\centering
\includegraphics[width=0.9\columnwidth]{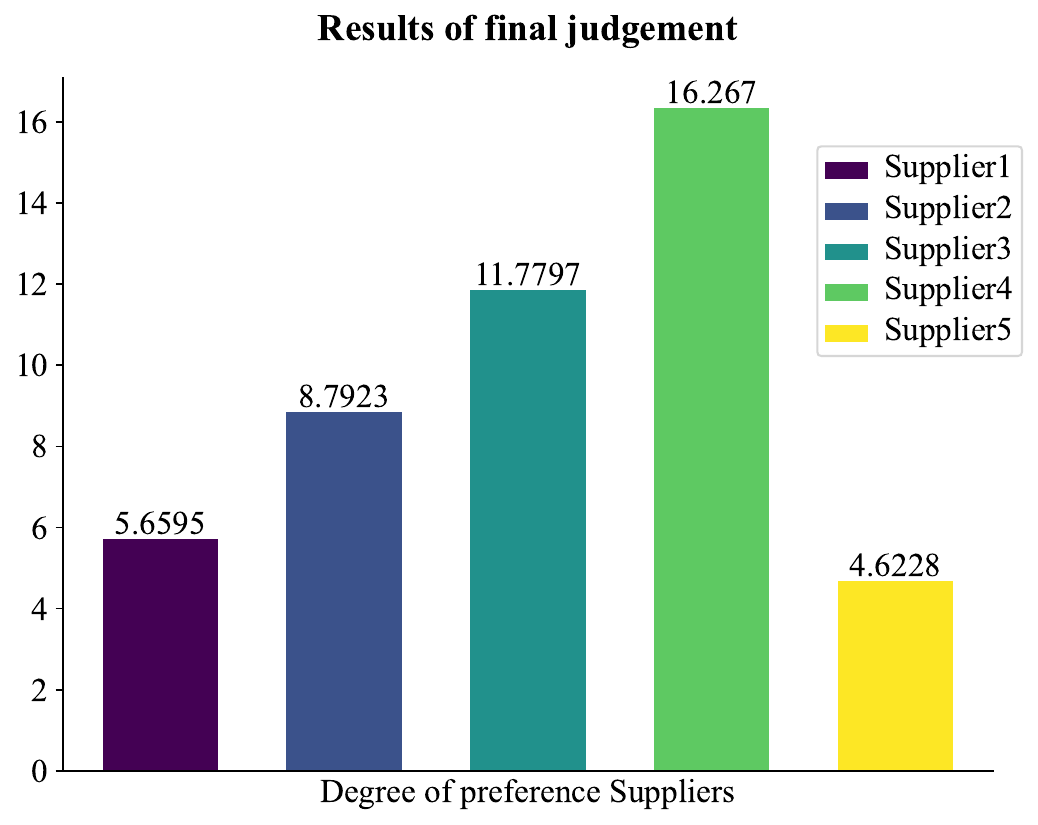}
\caption{Detailed process of novel framework of soft likelihood function}
\label{fig_2}
\end{figure}

In the round one, it can be concluded that the sequence of desired suppliers is $Supplier_3 > Supplier_4 > Supplier_2 > Supplier_1 > Supplier_5$, which is generally consistent with intuitive judgments. For example, by analyzing the information about $Supplier_3$, the modified affirmative part is bigger than other ones owned by information belongs to other suppliers,
\cite{ref23} which can be verified by checking the values obtained by improved soft likelihood function. Due to the effectiveness of soft likelihood function in information fusion , it is reasonable to regard the generated results are
intuitive and rational.

\subsection{Round 2}
In the second round, some necessary and crucial processed information is provided. The normalized degree of credibility, modified values of information value, attitude characters and the final judgments on suppliers are given in Table \ref{AE0.24},\ref{AE0.33},\ref{AE0.24},\ref{AD4.7} and Figure \ref{fig_2}, \ref{fig_3} respectively.

\begin{table}[ht]
\renewcommand{\arraystretch}{1.2}
\setlength{\tabcolsep}{14pt}
\centering
\begin{tabular}{c c c c}
\hline
Alternative  &$Expert_1$ &$Expert_2$&$Expert_3$ \\
\hline
$Supplier_1$ &0.2430 &0.2819 &0.4751\\
$Supplier_2$ &0.1859 &0.2734 &0.5407 \\
$Supplier_4$ &0.2430 &0.6099 &0.1472 \\
$Supplier_5$ &0.3423 &0.3466 &0.3111 \\
\hline
\end{tabular}
\caption{Degree of credibility of groups of IFSs}
\label{AE0.24}
\end{table}
\vspace{-0.3em}

\begin{table}[ht]
\renewcommand{\arraystretch}{1.2}
\setlength{\tabcolsep}{14pt}
\centering
\begin{tabular}{c c c c}
\hline
Alternative  &$Expert_1$ &$Expert_2$&$Expert_3$ \\
\hline
$Supplier_1$ &0.3382 &0.3287 &0.3331\\
$Supplier_2$ &0.3440 &0.3272 &0.3288 \\
$Supplier_4$ &0.3286 &0.3400 &0.3314 \\
$Supplier_5$ &0.3253 &0.3356 &0.3391 \\
\hline
\end{tabular}
\caption{Degree of credibility of groups of IFSs}
\label{AE0.33}
\end{table}

\begin{table}[ht]
\renewcommand{\arraystretch}{1.2}
\setlength{\tabcolsep}{14pt}
\centering
\begin{tabular}{c c c c}
\hline
Alternative  &$Expert_1$ &$Expert_2$&$Expert_3$ \\
\hline
$Supplier_1$ &0.2467 &0.2782 &0.4751\\
$Supplier_2$ &0.1931 &0.2701 &0.5368 \\
$Supplier_4$ &0.2376 &0.6172 &0.1452 \\
$Supplier_5$ &0.3342 &0.3469 &0.3166 \\
\hline
\end{tabular}
\caption{Attitude characters with respect to suppliers}
\label{AE0.24}
\end{table}

\begin{table}[htb]
\renewcommand{\arraystretch}{1.2}
\setlength{\tabcolsep}{15pt}
\centering
\begin{tabular}{c c c c}
\hline
&Alternative  &Degree of preference &\\
\hline
&$Supplier_1$ &4.770726973598357 &\\
&$Supplier_2$ &6.605666698011237 &\\
&$Supplier_4$ &14.521609619624286 &\\
&$Supplier_5$ &4.8011760761178754 &\\
\hline
\multicolumn{4}{c}{$Supplier_4>Supplier_2>Supplier_5>Supplier_1$}\\
\hline
\end{tabular}
\caption{Gross estimation with respect to suppliers}
\label{AD4.7}
\end{table}

\begin{figure}[htb]
\centering
\includegraphics[width=0.8\columnwidth]{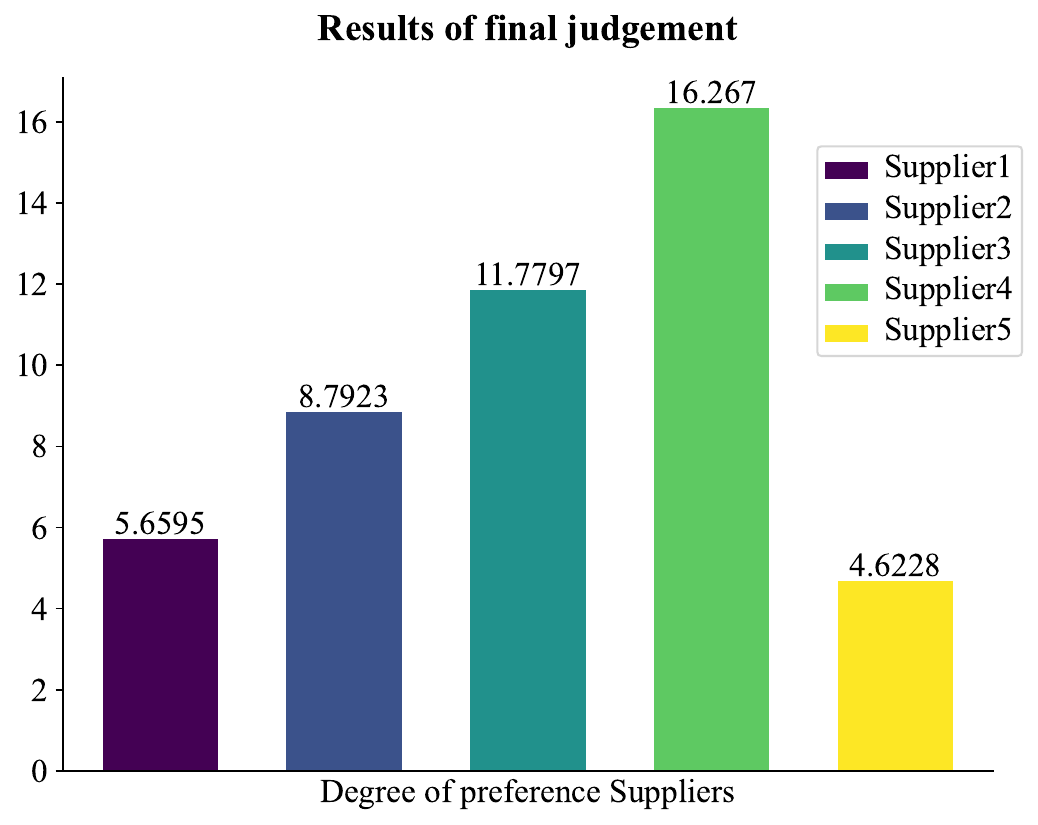}
\caption{Detailed process of novel framework of soft likelihood function}
\label{fig_2}
\end{figure}

\begin{figure}[htb]
\centering
\includegraphics[width=1\columnwidth]{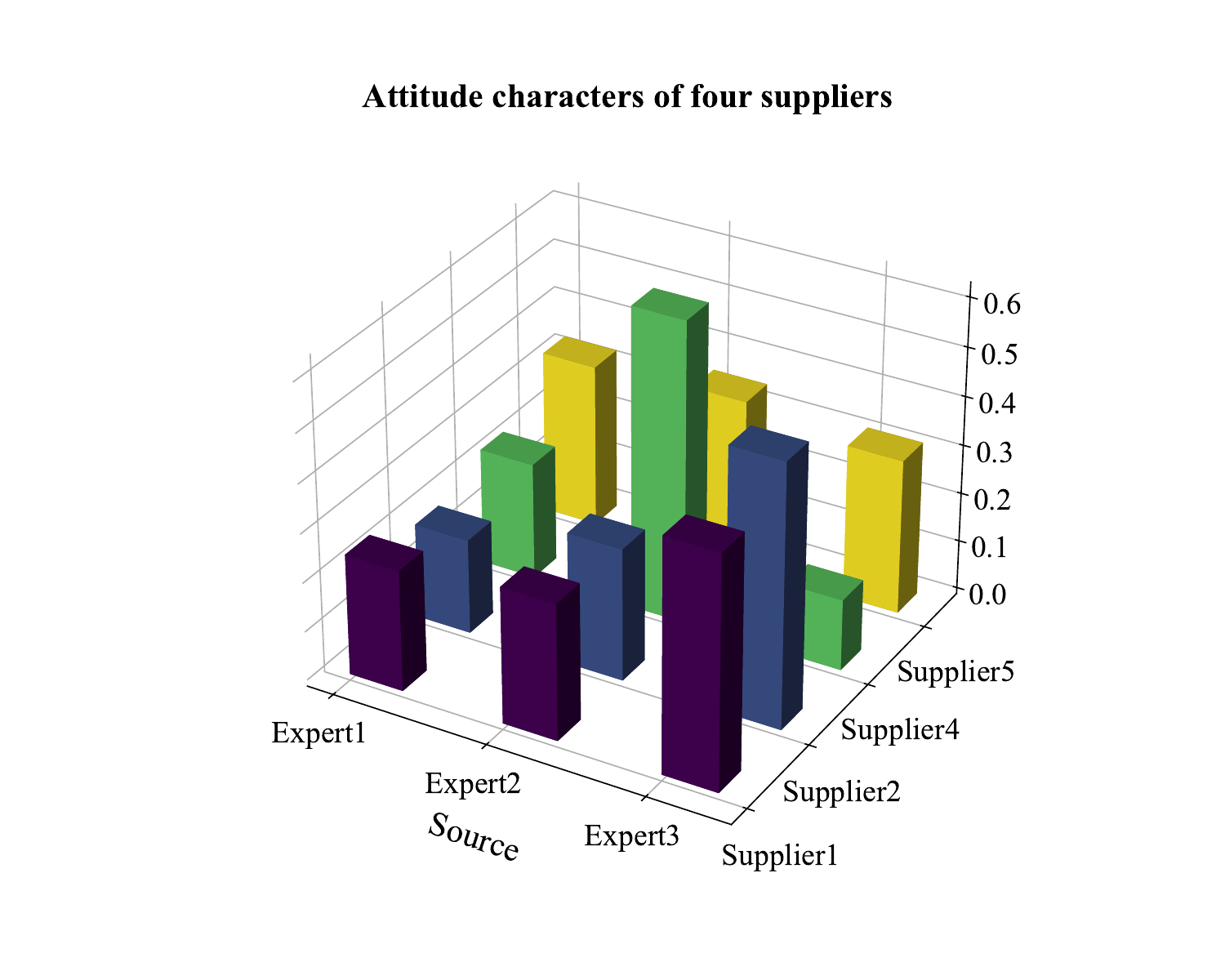}
\caption{Detailed process of novel framework of soft likelihood function}
\label{fig_3}
\end{figure}
In the second round, the level of priority of suppliers is $Supplier_4>Supplier_2>Supplier_5>Supplier_1$.

\subsection{ Round 3}
In the third round, some necessary and crucial processed information is provided. The normalized degree of credibility, modified values of information value and the final judgments on suppliers are given in Table \ref{A0.32}, \ref{A0.29}, \ref{A0.28} ,\ref{AD4.42} and Figure \ref{fig_3},\ref{fig_2} respectively.

In the second round, the level of priority of suppliers is $Supplier_6>Supplier_4>Supplier_3>Supplier_5>Supplier_2>Supplier_1$.

\begin{table}[htb]
\renewcommand{\arraystretch}{1.2}
\setlength{\tabcolsep}{14pt}
\centering

\begin{tabular}{c c c c}
\hline
Alternative  &$Expert_1$ &$Expert_2$&$Expert_3$ \\
\hline
$Supplier_1$ &0.3252 &0.3464 &0.3284\\
$Supplier_2$ &0.3379 &0.3401 &0.3220 \\
$Supplier_3$ &0.3345 &0.3238 &0.3417 \\
$Supplier_4$ &0.3260 &0.3430 &0.3310 \\
$Supplier_5$ &0.3348 &0.3184 &0.3468 \\
$Supplier_6$ &0.3458 &0.3459 &0.3083 \\
\hline
\end{tabular}
\caption{Degree of credibility of groups of IFSs}
\label{A0.32}
\end{table}

\begin{table}[H]
\renewcommand{\arraystretch}{1.2}
\setlength{\tabcolsep}{14pt}
\centering

\begin{tabular}{c c c c}
\hline
Alternative  &$Expert_1$ &$Expert_2$&$Expert_3$ \\
\hline
$Supplier_1$ &0.2949 &0.4356 &0.2694\\
$Supplier_2$ &0.3067 &0.3261 &0.3672 \\
$Supplier_3$ &0.3307 &0.2615 &0.4079 \\
$Supplier_4$ &0.3234 &0.4006 &0.2760 \\
$Supplier_5$ &0.3058 &0.4156 &0.2786 \\
$Supplier_6$ &0.1868 &0.5016 &0.3115 \\
\hline
\end{tabular}
\caption{Processed information volume of each group of IFSs}
\label{A0.29}
\end{table}

\begin{table}[H]
\renewcommand{\arraystretch}{1.2}
\setlength{\tabcolsep}{12pt}
\centering

\begin{tabular}{c c c c}
\hline
Alternative  &$Expert_1$ &$Expert_2$&$Expert_3$ \\
\hline
$Supplier_1$ &0.2860 &0.4501 &0.2639\\
$Supplier_2$ &0.3114 &0.3333 &0.3553 \\
$Supplier_3$ &0.3305 &0.2530 &0.4165 \\
$Supplier_4$ &0.3155 &0.4112 &0.2734 \\
$Supplier_5$ &0.3090 &0.3994 &0.2916 \\
$Supplier_6$ &0.1933 &0.5193 &0.2874 \\
\hline
\end{tabular}
\caption{Processed information volume of each group of IFSs}
\label{A0.28}
\end{table}

\begin{center}
\begin{table}[htb]
\renewcommand{\arraystretch}{1.3}
\setlength{\tabcolsep}{25pt}
\centering

\begin{tabular}{cc}
\hline
Alternative  &Degree of preference \\
\hline
$Supplier_1$ &4.424209272010652 \\
$Supplier_2$ &7.010993730725694 \\
$Supplier_3$ &14.140671975325015 \\
$Supplier_4$ &16.388871792488427 \\
$Supplier_5$ &13.152765205135305 \\
$Supplier_6$ &18.489995324457407 \\
\hline
\multicolumn{2}{c}{\makecell[c]{$Supplier_6>Supplier_4>Supplier_3>$\\$Supplier_5>Supplier_2>Supplier_1$}}\\
\hline
\end{tabular}
\caption{Gross estimation with respect to suppliers}
\label{AD4.42}
\end{table}
\end{center}

\begin{figure}[htb]
\centering
\includegraphics[width=0.8\columnwidth]{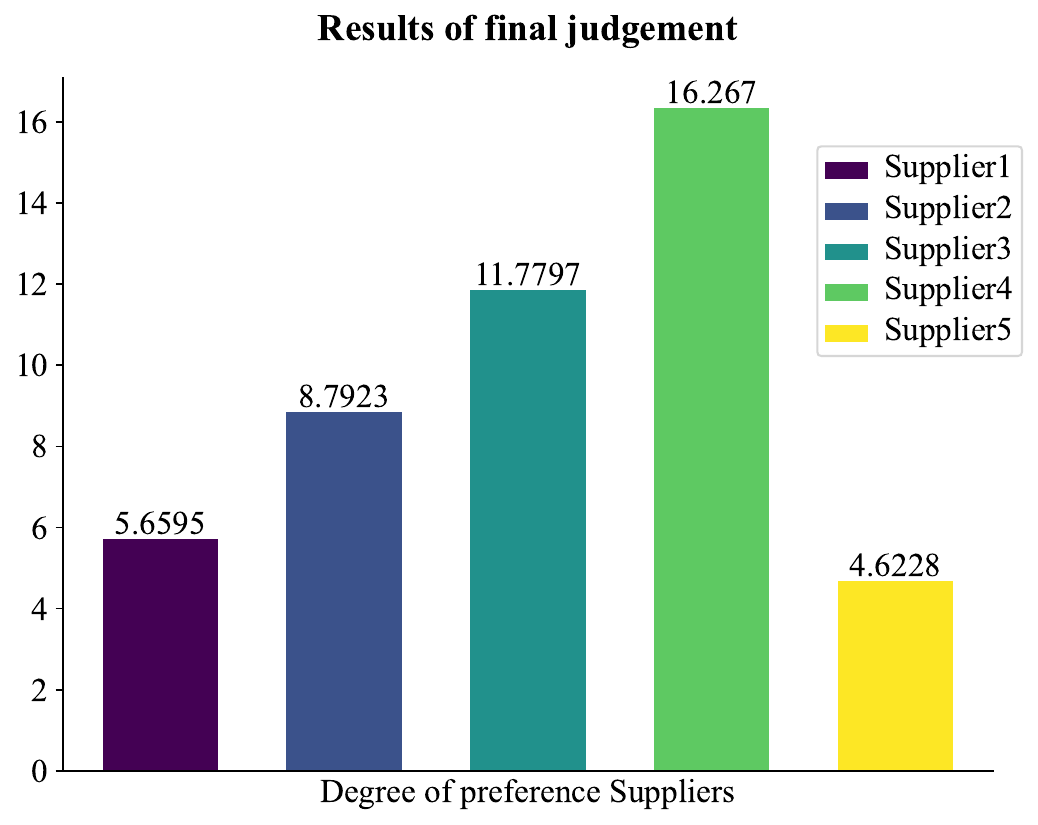}
\caption{Detailed process of novel framework of soft likelihood function}
\label{fig_2}
\end{figure}

\begin{figure}[htb]
\centering
\includegraphics[width=1\columnwidth]{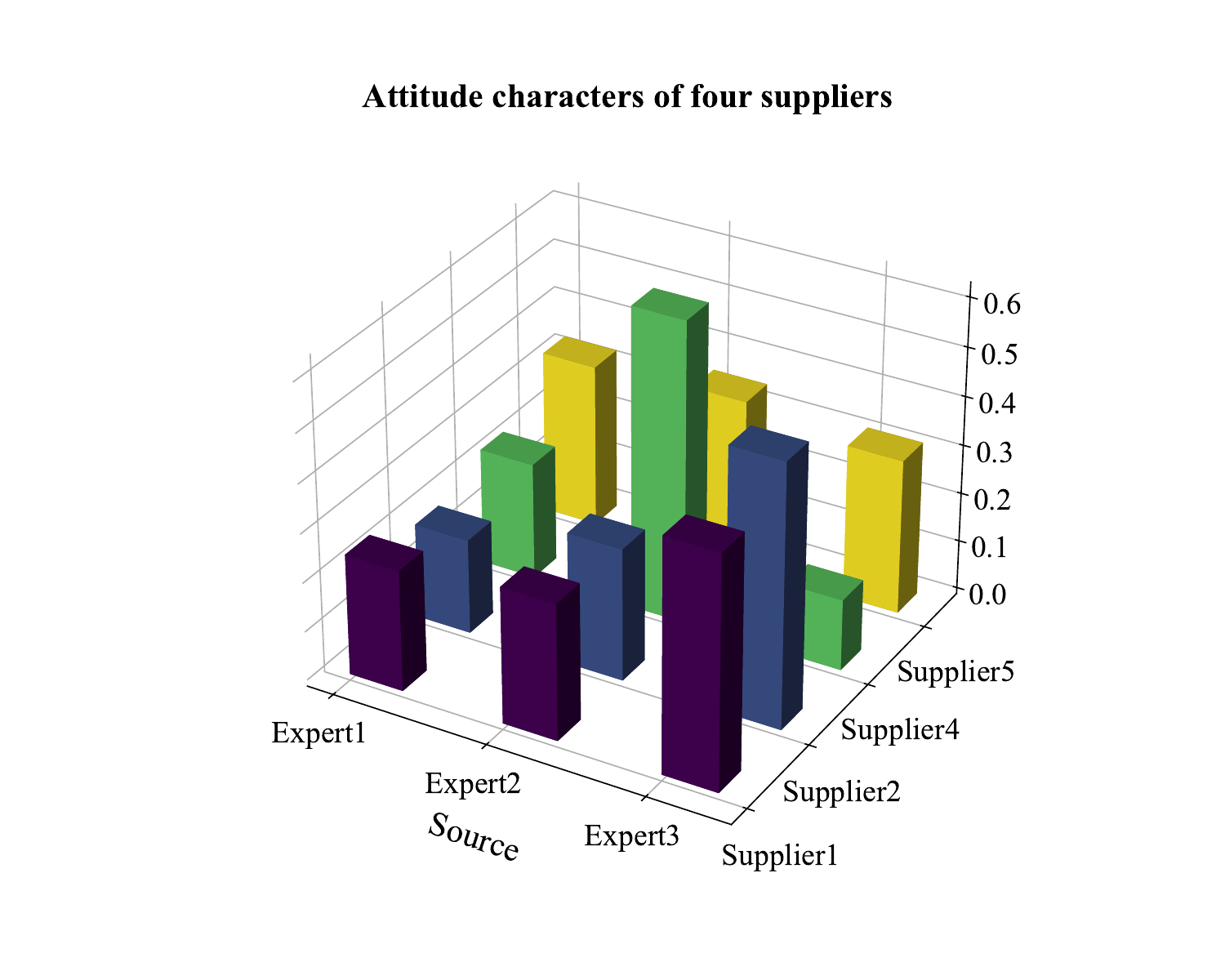}
\caption{Detailed process of novel framework of soft likelihood function}
\label{fig_3}
\end{figure}

\section{Comparison and discussions on the results produced by proposed method}
In this part, some comparisons are provided to verify the effectiveness of the proposed method and the results are provided in Table \ref{result}.

\subsection{Round 1}
In the first round, the results obtained by $Static MCDM$, $Dynamic MCDM$ and the proposed method are exactly the same. However, in regard to
AQM \cite{ref38}, the estimation about $Supplier_3$ and $Supplier_4$ are opposite compared with results of other three methods. The reasons for preferring $Supplier_4$ instead of $Supplier_3$ are well illustrated in \cite{ref39} from its own dimension of viewpoints. And the detailed causes are further explained on the base of data generated in the course of completing the process of the novel framework of soft likelihood function.

The first reason for this phenomenon is that the values of reliability measurement of $Supplier_4$ are generally bigger than that of $Supplier_3$, which
illustrates that $Supplier_4$ possesses more affirmative part of information according to the definition of reliability function. It indicates that experts mainly prefer $Supplier_4$ to $Supplier_3$ well.

The second reason is that the degree of credibility of $Supplier_4$ is closer to each other than $Supplier_3$, which demonstrates that the judgments from experts are consistent with each other and can be fully trusted. On the contrary, the information of $Supplier_4$ has more conflicting parts than $Supplier_3$, which can be obtained from the divergence of the value of degree of credibility. Therefore, when information conflicts, it is not convincing to distribute a comparatively high level of belief to it, which is fully embodied by the process of soft likelihood function \cite{ref23}.

The third reason is an extension of the second one. The information volume of $Supplier_3$ diverges much more than $Supplier_4$, which is also an indicator that judgments given by experts are very different from each other.Therefore, the information about Supplier3 transfers more negative signals
in the role of decision and strategy making.

In one word, the relation $S_4 > S_3$ is correct and valid.

\subsection{Round 2}
All of the methods reach an agreement in the judgments with respect to all of the Supplier discussed. Therefore, there is no need to have further
discussions about the final results of judgments.

\subsection{Round 3}
In the third round, it can be summarized that $AQM$, $Dynamic$ $MCDM$ and the proposed method reach an agreement on the estimation of actual situations. Moreover, some detailed causes are clearly given in \cite{ref39} based on its own techniques. And some other standpoints are provided utilizing data obtained by proposed method.

First, the degree of credibility of $Supplier_5$ are much more divergent than $Supplier_3$, which indicates that there exist conflicting information in the judgments about $Supplier_5$. Besides, it is also an evidence that the situation of information with respect to $Supplier_5$ are in chaos and the confidence level of it is lowered which embodies in the priority ranking of $Supplier_5$.

Second, the value of information volume is also a proof of the conclusion of the first part of reason. The information volume of $Supplier_3$ is
much more consistent than the one of $Supplier_5$, which proves that the information provided by experts may be extremely varied when compared
with each other.

Third, when checking the degree of reliability generated for $Supplier_3$ and $Supplier_5$ which is not provided, the mean value of the reliability of
$Supplier_3$ is much higher than the one of $Supplier_5$, which well illustrates that there exist a relatively bigger part of positive information about $Supplier_3$ according to the definition of reliability function.

All in all, the relation $S_3 > S_5$ is reasonable and rational.

\section{Conclusion}
In this paper, a novel framework of soft likelihood function is proposed
to serve for expert decision systems based on credibility measure, information volume and certain transformations. The proposed method fully
takes all factors which may have potential effects on the process of decision making, which ensures that correct judgments are accurately detected
and extracted. The final modified results well illustrates that the proposed
method possesses very excellent performance in managing multi-source
uncertain information and is able to handle productions of precise decisions under complex environments. All in all, the proposed method can be
regarded as a good solution in solving problems occurred in expert decision systems.


\bibliography{ref}

 



\end{document}